\documentclass{article}

\usepackage{microtype}
\usepackage{graphicx}
\usepackage{booktabs} 
\usepackage{algorithm}
\usepackage{algorithmic}
\usepackage{float}
\usepackage{amstext,amsmath,amssymb}
\usepackage{url}
\usepackage{booktabs}       
\usepackage{amsfonts}      
\usepackage{nicefrac}      
\usepackage{microtype}     
\usepackage{subfigure}
\usepackage{stfloats}

\usepackage{hyperref}

\usepackage[accepted]{icml2021}

\icmltitlerunning{The Emergence of Individuality}

\begin{document}

\twocolumn[
\icmltitle{The Emergence of Individuality}




\begin{icmlauthorlist}

\icmlauthor{Jiechuan Jiang}{pku}
\icmlauthor{Zongqing Lu}{pku}
\end{icmlauthorlist}

\icmlaffiliation{pku}{Peking University}
\icmlcorrespondingauthor{Zongqing Lu}{zongqing.lu@pku.edu.cn}

\icmlkeywords{Machine Learning, ICML}

\vskip 0.3in
]



\printAffiliationsAndNotice{}  

\begin{abstract}
	
	Individuality is essential in human society. It induces the division of labor and thus improves the efficiency and productivity. Similarly, it should also be a key to multi-agent cooperation. Inspired by that individuality is of being an individual separate from others, we propose a simple yet efficient method for the emergence of individuality (EOI) in multi-agent reinforcement learning (MARL). EOI learns a probabilistic classifier that predicts a probability distribution over agents given their observation and gives each agent an intrinsic reward of being correctly predicted by the classifier. The intrinsic reward encourages the agents to visit their own familiar observations, and learning the classifier by such observations makes the intrinsic reward signals stronger and in turn makes the agents more identifiable. To further enhance the intrinsic reward and promote the emergence of individuality, two regularizers are proposed to increase the discriminability of the classifier. We implement EOI on top of popular MARL algorithms. Empirically, we show that EOI outperforms existing methods in a variety of multi-agent cooperative scenarios. 
	
\end{abstract}

\section{Introduction}
Humans develop into distinct individuals due to both genes and environments \cite{freund2013emergence}. Individuality induces the division of labor \cite{gordon1996organization}, which improves the productivity and efficiency of human society. Analogically, the emergence of individuality should also be essential for multi-agent cooperation.

Although multi-agent reinforcement learning (MARL) has been applied to multi-agent cooperation, it is widely observed that agents usually learn similar behaviors, especially when the agents are homogeneous with shared global reward and co-trained \cite{mckee2020social}. For example, in multi-camera multi-object tracking \cite{liu2017multi}, where camera agents learn to cooperatively track multiple objects, the camera agents all tend to track the easy object. However, such similar behaviors can easily make the learned policies fall into local optimum. If the agents can respectively track different objects, they are more likely to solve the task optimally. Many studies formulate such a problem as task allocation or role assignment \cite{sander2002scalable, dastani2003role, sims2008automated}. However, they require that the agent roles are rule-based and the tasks are pre-defined, and thus are not general methods. Some studies intentionally pursue difference in agent policies by diversity \cite{Lee2020Learning, yang2019hierarchical} or by emergent roles \cite{wang2020multi}, however, the induced difference is not appropriately linked to the success of task. On the contrary, the emergence of individuality along with learning cooperation can automatically drive agents to behave differently and take a variety of roles, if needed, to successfully complete tasks.  

Biologically, the emergence of individuality is attributed to innate characteristics and experiences. However, as in practice RL agents are mostly homogeneous, we mainly focus on enabling agents to develop individuality through interactions with the environment during policy learning. Intuitively, in multi-agent environments where agents respectively explore and interact with the environment, individuality should emerge from what they experience. In this paper, we propose a novel method for the emergence of individuality (EOI) in MARL. EOI learns a probabilistic classifier that predicts a probability distribution over agents given their observation and gives each agent an intrinsic reward of being correctly predicted probability by the classifier. Encouraged by the intrinsic reward, agents tend to visit their own familiar observations. Learning the probabilistic classifier by such observations makes the intrinsic reward signals stronger and in turn makes the agents more identifiable. In this closed loop with positive feedback, agent individuality emerges gradually. However, at early learning stage, the observations visited by different agents cannot be easily distinguished by the classifier, meaning the intrinsic reward signals are not strong enough to induce agent characteristics. Therefore, we propose two regularizers for learning the classifier to increase the discriminability, enhance the feedback, and thus promote the emergence of individuality.

EOI is compatible with centralized training and decentralized execution (CTDE) methods. We realize EOI on top of two popular MARL methods, MAAC \cite{iqbal2019actor} and QMIX \cite{rashid2018qmix}. For MAAC, as each agent has its own critic, it is convenient to shape the reward for each agent. For QMIX, we introduce an auxiliary gradient and update the individual value function by both minimizing the TD error of the joint action-value function and maximizing its cumulative intrinsic rewards. In experiments, we verify the effectiveness of the intrinsic reward and confirm that the proposed regularizers indeed improve the emergence of individuality even if agents have the same innate characteristics by ablation studies. And we empirically demonstrate that EOI outperforms existing methods in both grid-world and large-scale environments. Finally, we discuss and numerically show the similarity and difference between EOI and DIAYN \cite{eysenbach2018diversity}.

\section{Related Work}

\subsection{MARL}

We consider the formulation of Decentralized Partially Observable Markov Decision Process (Dec-POMDP) \cite{oliehoek2016concise}. The are $n$ agents in an environment. At each timestep $t$ each agent $i$ receives a local observation $o^t_i$, takes an action $a^t_i$, and gets a shared global reward $r^t$. Agents together aim to maximize the expected return $\mathbb{E}\sum_{t=0}^{T}\gamma ^t r^t$, where $\gamma$ is a discount factor and $T$ is the time horizon. Many methods have been proposed for Dec-POMDP, most of which follow the paradigm of centralized training and decentralized execution. Value function factorization methods decompose the joint value function into individual value functions. VDN \cite{sunehag2018value} and QMIX \cite{rashid2018qmix} respectively propose additivity and monotonicity for factorization structures. Qatten \cite{yang2020qatten} is a variant of VDN, which uses a multi-head attention structure to utilize global information. QPLEX \cite{wang2021qplex} uses a duplex dueling network architecture for factorization and theoretical analyzes the full representation expressiveness. Some methods extend policy gradient into multi-agent cases, which contain a centralized critic with global information and decentralized actors which only have access to local information. COMA \cite{foerster2018counterfactual} proposes a counterfactual baseline for multi-agent credit assignment. MADDPG \cite{lowe2017multi} is an extension of DDPG algorithm \cite{lillicrap2016continuous}. DOP \cite{wang2021dop} replaces the conventional critic with a value decomposed critic. Communication methods \cite{das2019tarmac,jiang2020graph,ding2020learning} share information between agents for implicit coordination, and DCG \cite{bohmer2020deep} adopts coordination graphs for explicit coordination.

\subsection{Behavior Diversification}

Many cooperative multi-agent applications require agents to take different behaviors to complete the task successfully. Behavior diversification can be handcrafted or emerge through agents' learning. Handcrafted diversification is widely studied as task allocation or role assignment. Heuristics \cite{sander2002scalable, dastani2003role, sims2008automated, macarthur2011distributed} assign specific tasks or pre-defined roles to each agent based on goal, capability, visibility, or by search. M$^3$RL \cite{shu2018mrl} learns a manager to assign suitable sub-tasks to rule-based workers with different preferences and skills. These methods require that the sub-tasks and roles are pre-defined, and the worker agents are rule-based. However, in general, the task cannot be easily decomposed even with domain knowledge and workers are learning agents.

The emergent diversification for single agent has been studied in DIAYN \cite{eysenbach2018diversity}, which learns reusable diverse skills in complex and transferable tasks without any reward signal by maximizing the mutual information between states and skill embeddings as well as entropy. In multi-agent learning, SVO \cite{mckee2020social} introduces diversity into heterogeneous agents for more generalized and high-performing policies in social dilemmas. Some methods are proposed for behavior diversification in multi-agent cooperation. ROMA \cite{wang2020multi} learns a role encoder to generate role embedding, and learns a role decoder to generate neural network parameters from embedding. Two regularizers are introduced for learning identifiable and specialized roles. However, due to large parameter space, generating various parameters for the emergent roles is inefficient. And mode collapse would happen in the role decoder even with different role embeddings. Learning low-level skills for each agent using DIAYN is considered in \cite{Lee2020Learning,yang2019hierarchical}, where agents’ diverse low-level skills are coordinated by the high-level policy. However, the diversity is not considered in the high-level policy. 

\section{Method}

Individuality is of being an individual separate from others. Motivated by this, we propose EOI, where agents are intrinsically rewarded in terms of being correctly predicted by a probabilistic classifier that is learned based on agents' observations. If the classifier learns to accurately distinguish agents, agents should behave differently and thus individuality emerges. Two regularizers are introduced for learning the classifier to enhance intrinsic reward signal and promote individuality. EOI directly correlates individuality with the task by intrinsic reward, and thus individuality emerges naturally during agents' learning. EOI can be applied to Dec-POMDP tasks and trained along with CTDE algorithms. We design practical techniques to implement EOI on top of two popular MARL methods, MAAC and QMIX.

\subsection{Intrinsic Reward}

\begin{figure}[t]
	\setlength{\abovecaptionskip}{3pt}
	\centering
	\includegraphics[width=.3\textwidth]{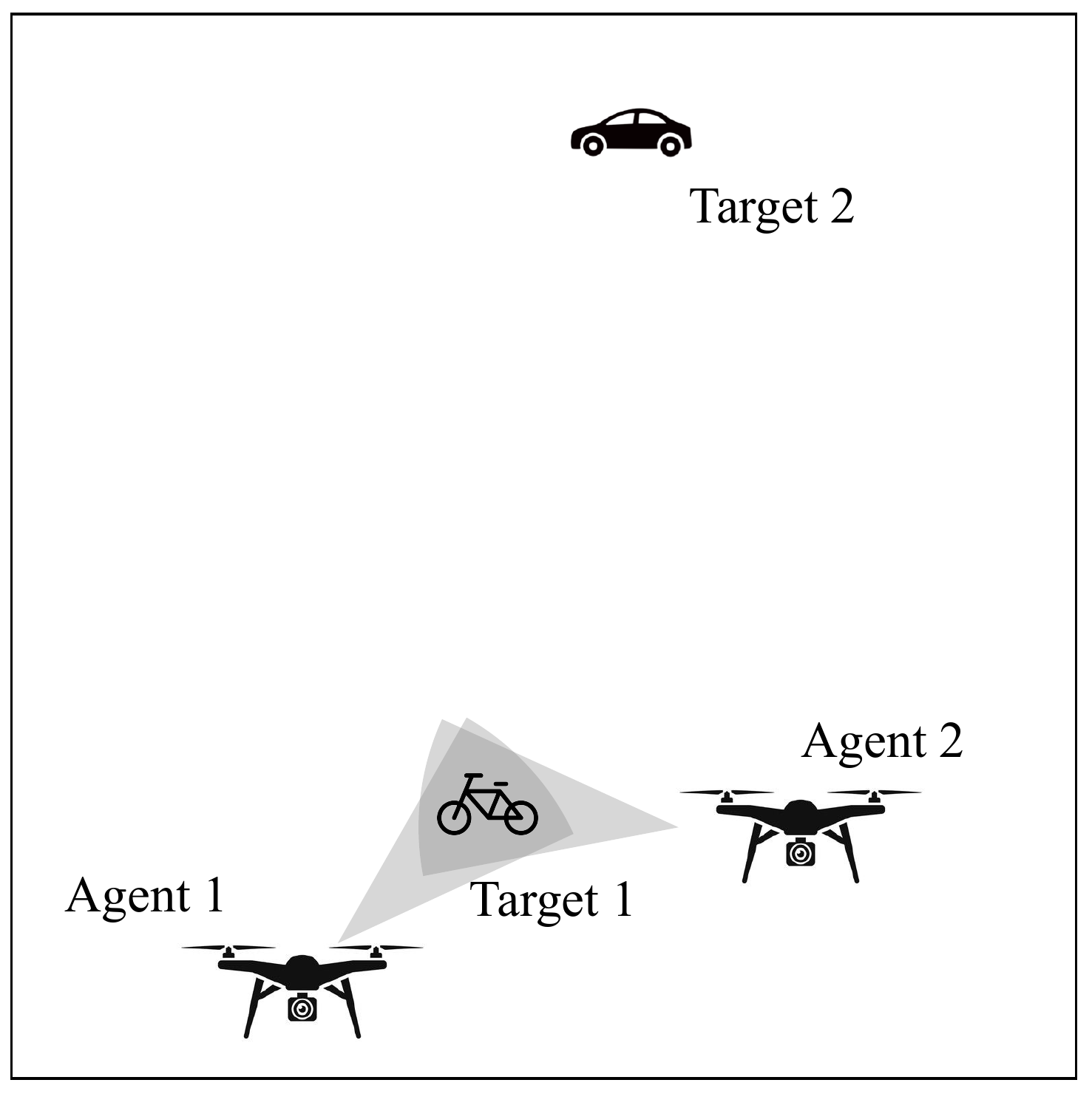}
	\vspace{-0.15cm}
	\caption{Multi-camera multi-target capturing}
	\label{fig:tracking}
\end{figure}

As illustrated in Figure~\ref{fig:tracking}, two camera agents learn to capture two targets, where the closer and slower target $1$ is the easier one. Capturing each target, they will get a global reward of $+1$. Sequentially capturing the two targets by one agent or together is sub-optimal, sometimes impossible when the task has a limited time horizon. The optimal solution is that the two agents go capturing different targets simultaneously. It is easy for both agents to learn to capture the easier target $1$. However, after that, target $2$ becomes even harder to be explored, and the learned policies easily fall at local optimum. Nevertheless, the emergence of individuality can address the problem, \emph{e.g.}, agents prefer to capture different targets.

To enable agents to develop individuality, EOI learns a probabilistic classifier $P(I|O)$ to predict a probability distribution over agents given on their observation, and each agent takes the correctly predicted probability as the intrinsic reward at each timestep. Thus, the reward function for agent $i$ is modified as
\begin{equation}
\nonumber 
r+\alpha p(i|o_i),
\end{equation}
where $r$ is the global environmental reward, $p(i|o_i)$ is the predicted probability of agent $i$ given its observation $o_i$, and $\alpha$ is a tuning hyperparameter to weight the intrinsic reward. With the reward shaping, EOI works as follows. If there is initial difference between agent policies in terms of visited observations, the difference is captured by $P(I|O)$ as it is fitted using agents' experiences. The difference is then fed back to each agent as an intrinsic reward. As agents maximize the expected return, the difference in agents' policies is exacerbated together with optimizing the environmental return. Therefore, the learning process is a closed loop with positive feedback. As agents progressively behave more identifiably, the classifier can distinguish agents more accurately, and thus individuality emerges gradually.

\begin{figure}[t]
	\setlength{\abovecaptionskip}{3pt}
	\centering
	\includegraphics[width=.4\textwidth]{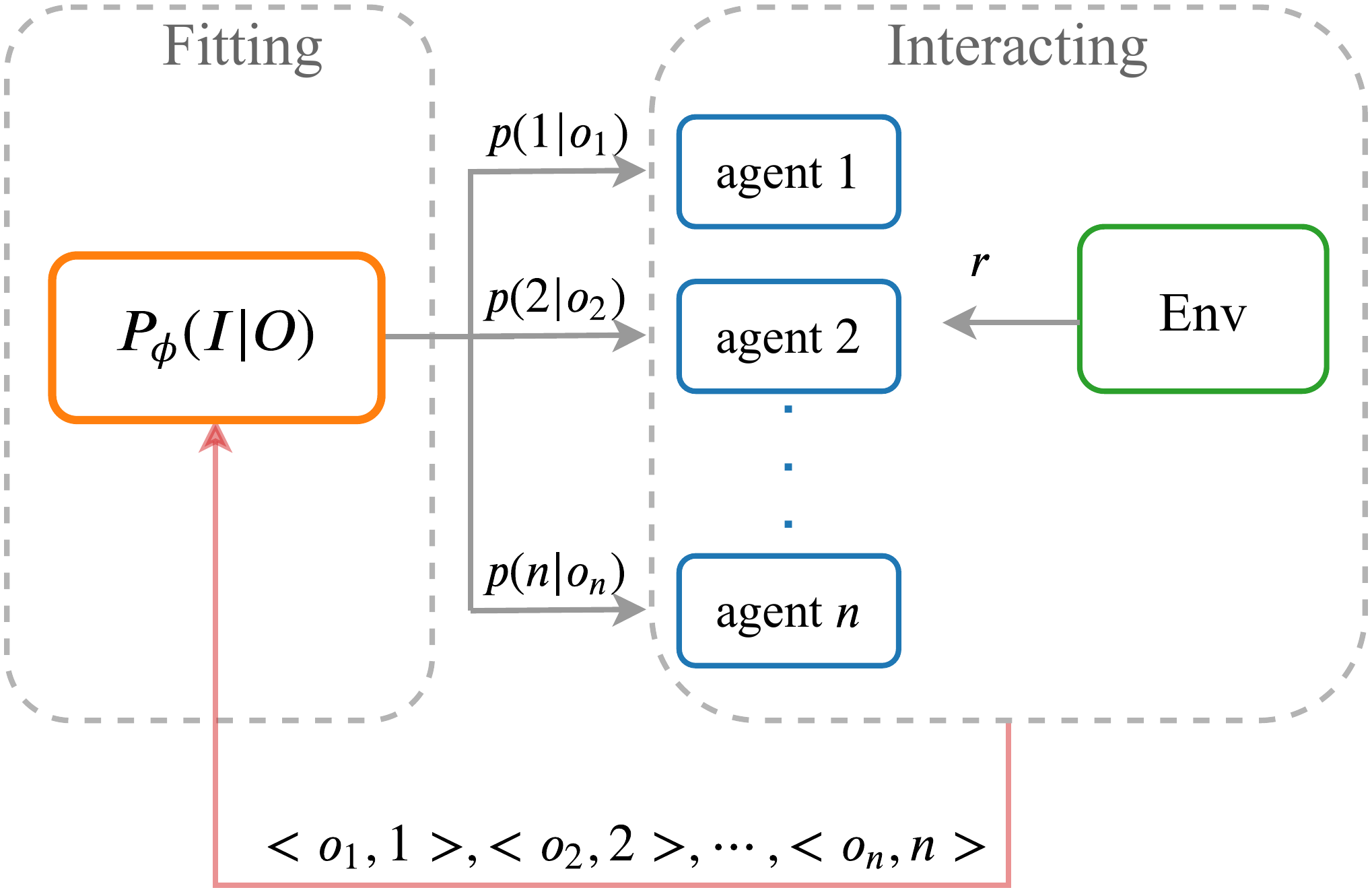}
	\vspace{-0.15cm}
	\caption{EOI}
	\label{fig:eoi}
\end{figure}

The classifier $P_\phi(I|O)$ is parameterized by a neural network $\phi$ and learned in a supervised way. At each timestep, we take each agent $i$'s observation $o_i$ as input and the agent index $i$ as the label and store the pair $<o_i,i>$ into a buffer $\mathcal{B}$. $\phi$ is updated by minimizing the cross-entropy loss ($\mathrm{CE}$), which is computed based on the uniformly sampled batches from $\mathcal{B}$. The learning process of EOI is illustrated in Figure~\ref{fig:eoi}.

\subsection{Regularizers of $P_{\phi}(I|O)$}
In the previous section, we assume that there is some difference between agents' policies. However, in general, the difference between initial policies is small (even no differences if agents' policies are initially by the same network weights), and the policies will quickly learn similar behaviors as in the example in Figure~\ref{fig:tracking}. Therefore, the intrinsic rewards are nearly the same for each agent, which means no feedback in the closed loop. To generate the feedback in the closed loop, the observation needs to be identifiable and thus the agent can be distinguished in terms of observations by $P_{\phi}(I|O)$. To address this, we propose two regularizers: positive distance (PD) and mutual information (MI) for learning $P_{\phi}(I|O)$.

\begin{figure*}[!h]
	\setlength{\abovecaptionskip}{3pt}
	\hspace{2.5cm}
	\subfigure[EOI+MAAC]
	{
		\setlength{\abovecaptionskip}{3pt}
		\includegraphics[width=0.3\textwidth]{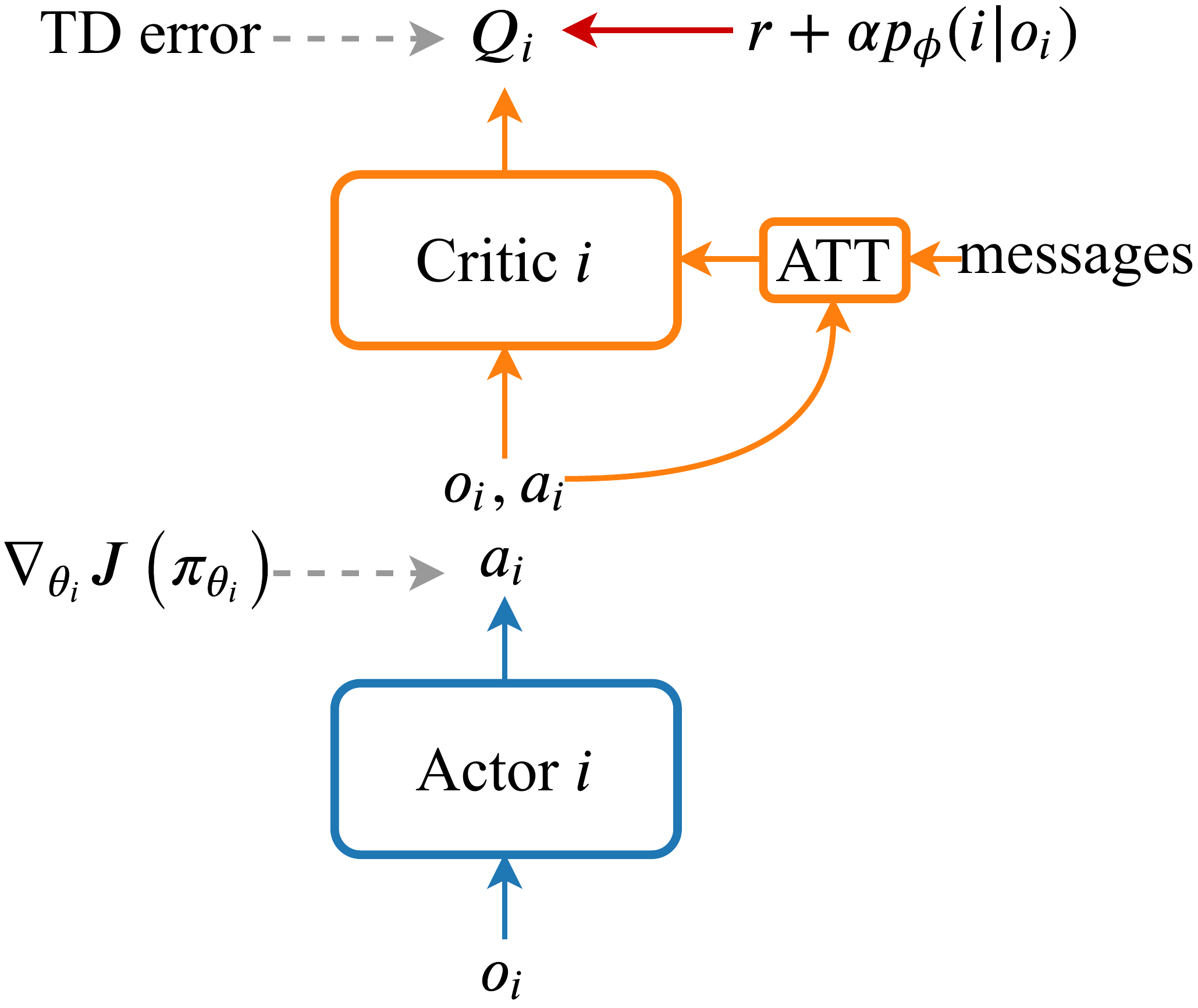}
		\label{fig:maac}
	}
	\hspace{0.5cm}
	\subfigure[EOI+QMIX]
	{
		\setlength{\abovecaptionskip}{3pt}
		\includegraphics[width=0.36\textwidth]{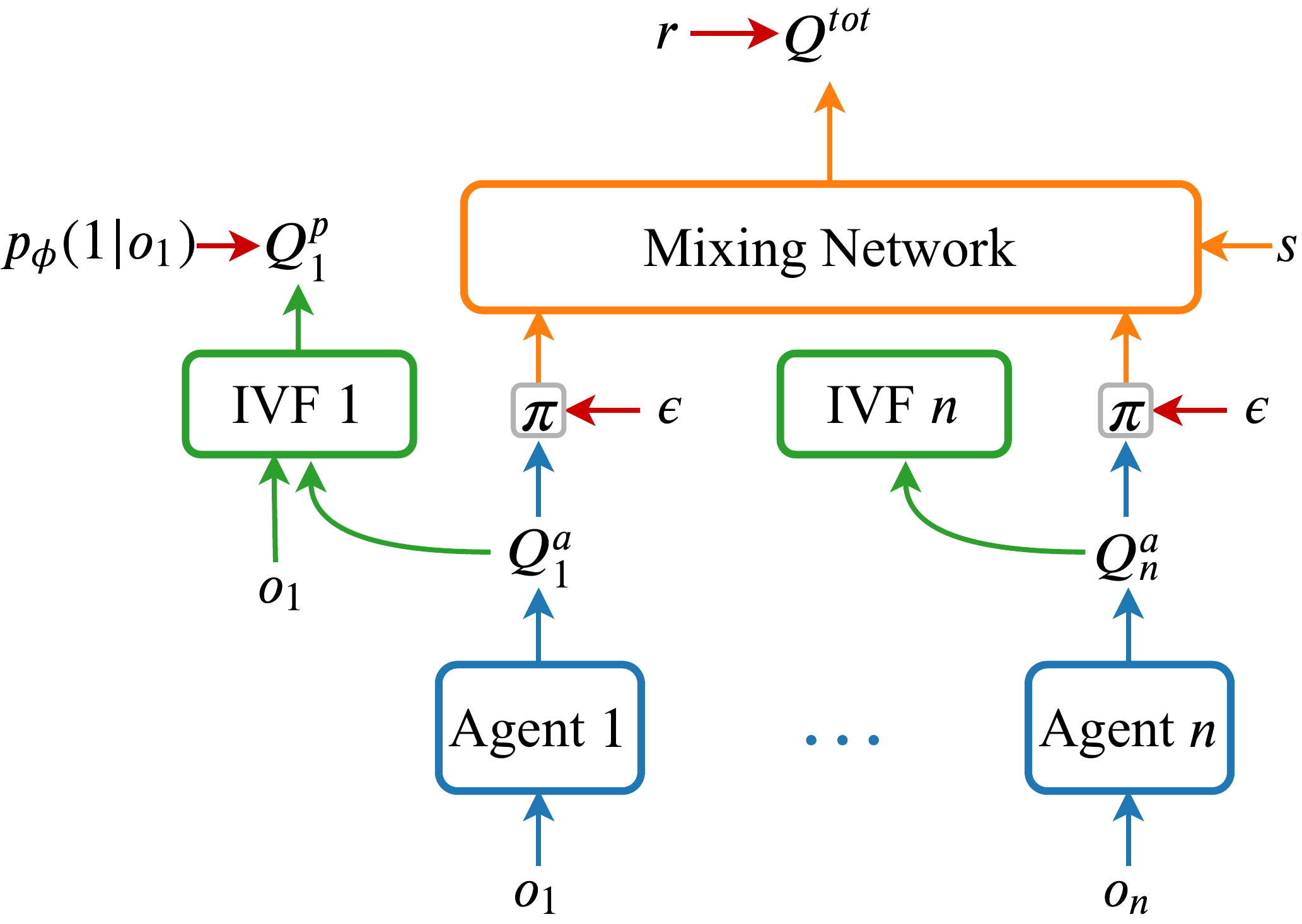}
		\label{fig:qmix}
	}
	\vspace{-0.20cm}
	\caption{Illustration of EOI with MAAC and QMIX.}
\end{figure*}

\textbf{Positive Distance.} The positive distance is inspired from the triplet loss \cite{schroff2015facenet} in contrastive learning, which is proposed to learn identifiable embeddings. Since $o^t_i$ and its previous observations $\{o_i^{t-\Delta t},o_i^{t-\Delta t+1},\cdots,o_i^{t-1}\}$ are distributed on the trajectory generated by agent $i$, the previous observations in the $\Delta t$-length window could be seen as the positives of $o_i^t$. To make the probability distribution on the anchor $o_i^t$ close to that on the positives, we sample an observation $o_i^{t'}$ from $\{o_i^{t-\Delta t},o_i^{t-\Delta t+1},\cdots,o_i^{t-1}\}$ and minimize the cross-entropy loss
\begin{equation}
\nonumber
\mathrm{CE}\left(p_{\phi}(\cdot|o_i^t),p(\cdot|o_i^{t'})\right).
\end{equation}
The positive distance minimizes the intra-distance between the observations with the same ``identity'', which hence enlarges the margin between different ``identities''. As a result, the observations become more identifiable. Since the positives are naturally defined on the trajectory, the identifiability generated by the positive distance is actually induced by the agent policy. Defining the negatives is hard but we find that just using the positive distance works well in practice.

\textbf{Mutual Information.} If the observations are more identifiable, it is easier to infer the agent that visits the given observation most, which indicates the higher mutual information between the agent index and observation. Therefore, to further increase the discriminability of the classifier, we maximize their mutual information,
\begin{equation}	 
\nonumber
\begin{split}
\mathrm{MI}(I;O) &= \mathcal{H}(I)-\mathcal{H}(I|O)\\
& = \mathcal{H}(I)-\mathbb{E}_{o\sim p(o)}\left [\sum_{i}-p(i|o)\log p(i|o)\right ].
\end{split}
\end{equation}
Since we store $<o_i,i>$ of every agent in $\mathcal{B}$, the number of samples for each agent is equal. Fitting $P_{\phi}(I|O)$ using batches from $\mathcal{B}$ ensures $\mathcal{H}(I)$ is a constant. To maximize $\mathrm{MI}(I;O)$ is to minimize $\mathcal{H}(I|O)$. Therefore, equivalently, we sample batches from $\mathcal{B}$ and minimize
\begin{equation}
\nonumber
\mathrm{CE}\left(p_{\phi}(\cdot|o_i^t),p_{\phi}(\cdot|o_i^{t})\right).
\end{equation}
Therefore, the optimization objective of $P_{\phi}(I|O)$ is to minimize
\begin{equation} 
\nonumber
\begin{gathered}
\mathrm{CE}\left(p_{\phi}(\cdot|o_i^t),\mathrm{one\_hot}(i)\right)+\beta_1 \mathrm{CE}\left(p_{\phi}(\cdot|o_i^t),p(\cdot|o_i^{t'})\right)\\
+\beta_2 \mathrm{CE}\left(p_{\phi}(\cdot|o_i^t),p_{\phi}(\cdot|o_i^{t})\right),
\end{gathered}
\end{equation}
where $\beta_1$ and $\beta_2$ are hyperparameters. The regularizers increase the discriminability of $P_{\phi}(I|O)$, make the intrinsic reward signals stronger to stimulate the agents to be more distinguishable, and eventually promote the emergence of individuality. In this sense, $P_{\phi}(I|O)$ not only is the posterior statistics, but also serves as the inductive bias of agents' learning.

\subsection{Implementation with MAAC and QMIX}

Existing methods for reward shaping in MARL focus on independent learning agents \cite{mckee2020social,Wang*2020Influence-Based,du2019liir}. How to shape the reward in Dec-POMDP for centralized training has not been deeply studied. Since the intrinsic reward can be exactly assigned to the specific agent, individually maximizing the intrinsic reward is more efficient than jointly maximizing the sum of all agents' intrinsic rewards \cite{hughes2018inequity}. Adopting this idea, we respectively present the implementation with MAAC and QMIX for realizing EOI. MAAC is an off-policy actor-critic algorithm, where each agent learns its own critic, thus it is convenient to directly give the shaped reward $r+\alpha p_{\phi}(i|o_i)$ to the critic of each agent $i$, without modifying other components, as illustrated in Figure~\ref{fig:maac}. The TD error of the critic and the policy gradient are the same as in MAAC \cite{iqbal2019actor}.

In QMIX, each agent $i$ has an individual action-value function $Q^a_i$. All the individual action-value functions are monotonically mixed into a joint action-value $Q^{tot}$ by a mixing network. Each agent selects the action with the highest individual value, but the individual value has neither actual meaning nor constraints \cite{rashid2018qmix}. Therefore, we can safely introduce an auxiliary gradient of the intrinsic reward to the individual action-value function $Q^a_i$. Each agent $i$ learns an intrinsic value function (IVF) $Q^p_i$, which takes as input the observation $o_i$ and the individual action-value vector $Q^a_i(o_i)$ and approximates $\mathbb{E}\sum_{t=0}^T \gamma^t p(i|o_i^t)$ by minimizing the TD error, 
\begin{equation}
\nonumber
\begin{gathered}
\mathbb{E}_{<o_i, Q^a_i(o_i), o_i^{\prime}> \sim \mathcal{D}}\left[(Q_i^p\left(o_i, Q^a_i(o_i)\right)-y)^{2}\right], \\
\text { where } y=p_{\phi}(i|o_i)+\gamma \bar{Q}^p_i(o'_i, \bar{Q}^a_i(o'_i)).
\end{gathered}
\end{equation}
$\bar{Q}_i^a$ and $\bar{Q}_i^p$ are the target value functions and $\mathcal{D}$ is the replay buffer. In order to improve both the global reward and intrinsic reward, we update $Q_i^a$, parameterized by $\theta_i$, towards maximizing $\mathbb{E}\left[Q^p_i\left(o_i, Q^a_i(o_i;\theta_i)\right)\right]$ along with minimizing the TD error of $Q^{tot}$ (denoted as $\delta^{tot}$), as illustrated in Figure~\ref{fig:qmix}. Since the intrinsic value function is differentiable with respect to the individual action-value vector $Q_i^a(o_i;\theta_i)$, and the action-value vector is continuous, we can establish the connection between $Q_i^p$ and $Q_i^a$ by the chain rule, like the policy update in DDPG \cite{lillicrap2016continuous}, and the gradient of $\theta_i$ is,
\begin{equation}
\nonumber
\nabla_{\theta_i}  J(\theta_i) = \frac{\partial \delta^{tot}}{\partial \theta_i} - \alpha\frac{\partial Q^p_i(o_i,Q^a_i(o_i;\theta_i))}{\partial \theta_i}.
\end{equation}

MAAC and QMIX are off-policy algorithms, and the environmental rewards are stored in the replay buffer. However, the intrinsic rewards are recomputed in the sampled batches before each update, since $P_{\phi}(I|O)$ is co-evolving with the learning agents and hence the previous intrinsic reward is outdated. 
The joint learning process of the classifier and agent policies can be mathematically formulated as a bi-level optimization, which is detailed in Appendix \ref{app:bilevel}. 
Note that in EOI, we can easily replace local observation with trajectory by taking the hidden state of RNN (it takes the trajectory as input) as the input of the classifier, which might be helpful in complex environments.

\begin{figure}[!t]
	\setlength{\abovecaptionskip}{3pt}
	\centering
	\subfigure[Pac-Men]
	{
		\setlength{\abovecaptionskip}{3pt}
		\includegraphics[width=0.22\textwidth]{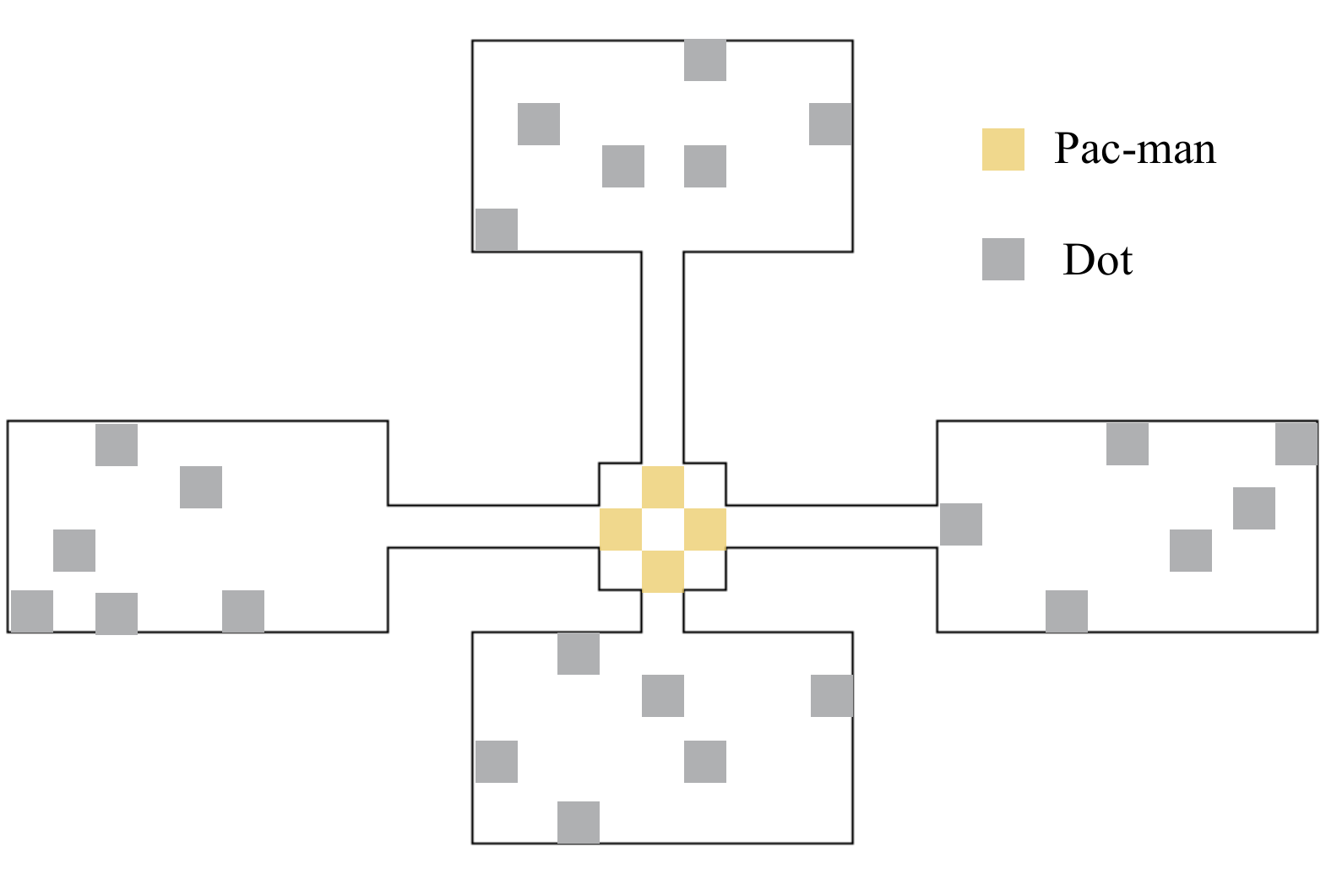}
	}
	\subfigure[Windy Maze ]
	{
		\setlength{\abovecaptionskip}{3pt}
		\includegraphics[width=0.22\textwidth]{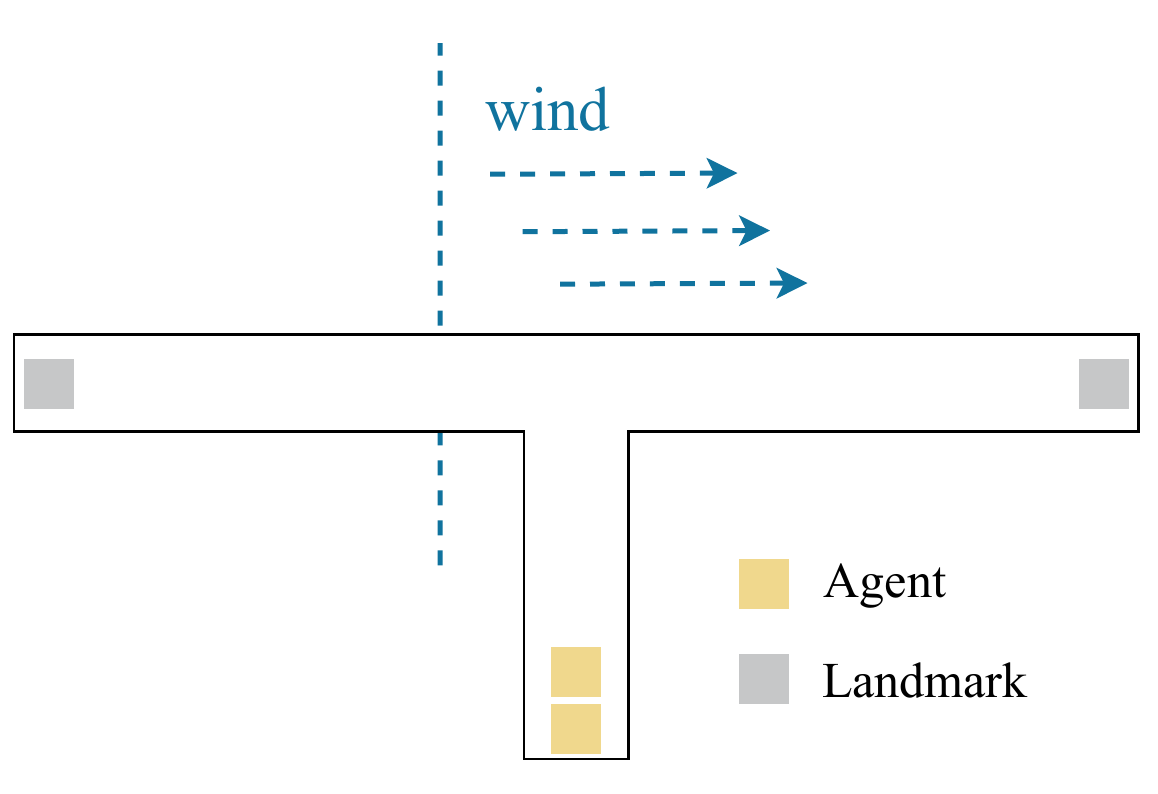}
	}
	\vspace*{-0.3cm}
	\caption{Illustration of Pac-Men and Windy Maze.}
	\label{fig:Illustration_1}
	\vspace{-0.1cm}
\end{figure}

\section{Experiments}

In the evaluation, we first verify the effectiveness of the intrinsic reward and the two regularizers on both MAAC and QMIX by ablation studies. Then, we compare EOI against EDTI \cite{Wang*2020Influence-Based} to verify the advantages of individuality with sparse reward, and against ROMA \cite{wang2020multi} and \citet{Lee2020Learning} (denoted as HC) to investigate the advantages of individuality over emergent roles and diversity. We also discuss the similarity and difference between EOI and DIAYN \cite{eysenbach2018diversity}, and provide the numerical results.
In grid-world environments, the agents do not share the weights of neural network since parameter sharing causes similar agent behaviors. All the curves are plotted using mean and standard deviation. The details about the experimental settings and the hyperparameters are available in Appendix \ref{app:hyper}. 

\subsection{Performance and Ablation}

To clearly interpret the mechanism of EOI, we test EOI in the two scenarios: Pac-Men and Windy Maze, as illustrated in Figure~\ref{fig:Illustration_1}.
\begin{itemize}
	\item \textit{Pac-Men.} There are four agents initialized at the maze center, and some randomly initialized dots. At each timestep, each agent can move to one of four neighboring grids or eat a dot. 
	\item \textit{Windy Maze.} There are two agents initialized at the bottom of the T-shaped maze, and two dots initialized at the two ends. At each timestep, each agent could move to one of four neighboring grids or eat a dot. There is a wind running right from the dotted line. Shifted by the wind, forty percent of the time, the agent will move to the right grid whichever action it takes. 
\end{itemize}
Each agent has a local observation that contains a square view with \(5\times5\) grids centered at the agent itself, and could only get a global reward, \emph{i.e.}, the total eaten dots, at the final timestep.

\begin{figure}[!t]
	\setlength{\abovecaptionskip}{3pt}
	\centering
	\subfigure[EOI+MAAC]
	{
		\setlength{\abovecaptionskip}{3pt}
		\includegraphics[width=0.23\textwidth]{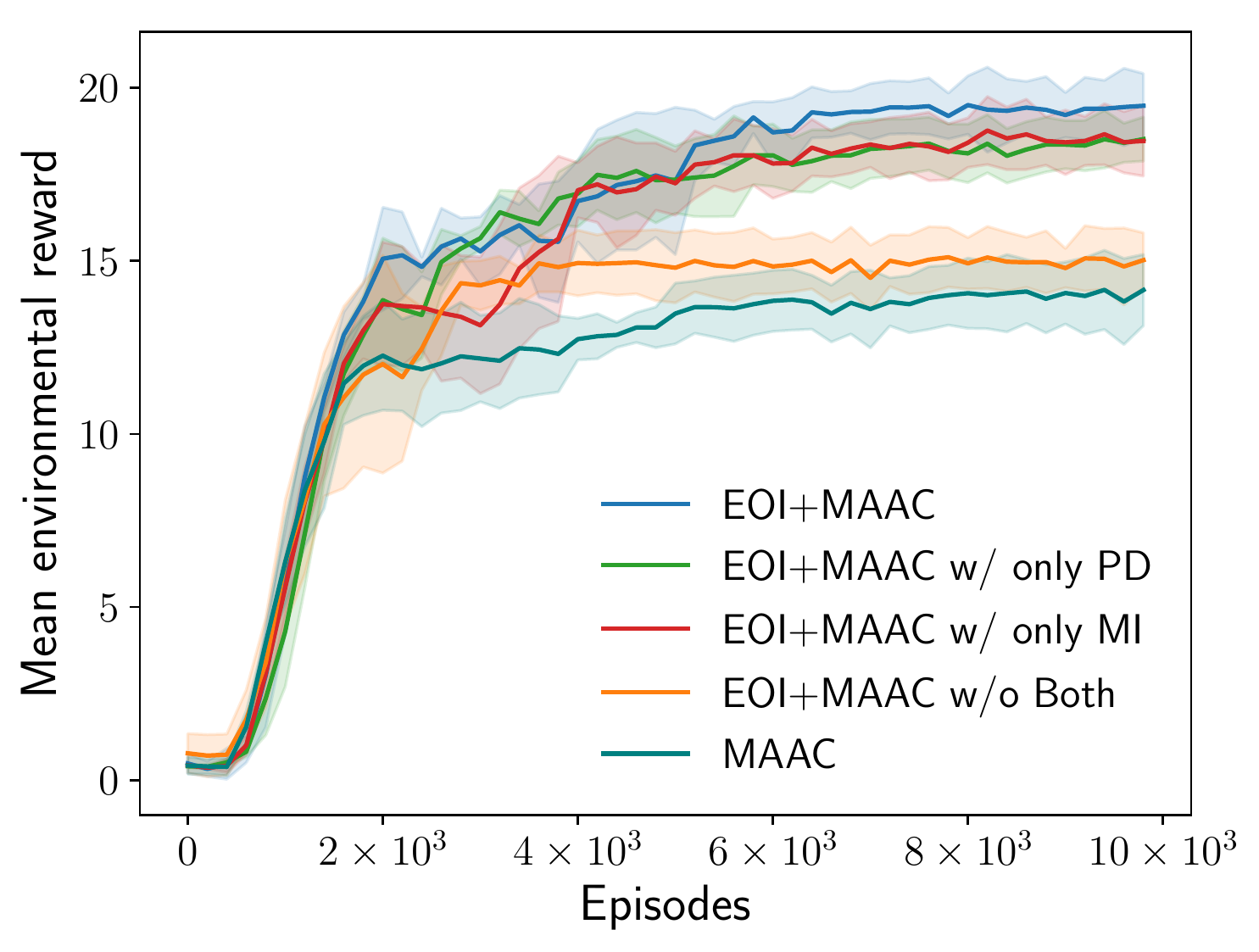}
	}
	\hspace{-0.60cm}
	\subfigure[EOI+QMIX]
	{
		\setlength{\abovecaptionskip}{3pt}
		\includegraphics[width=0.23\textwidth]{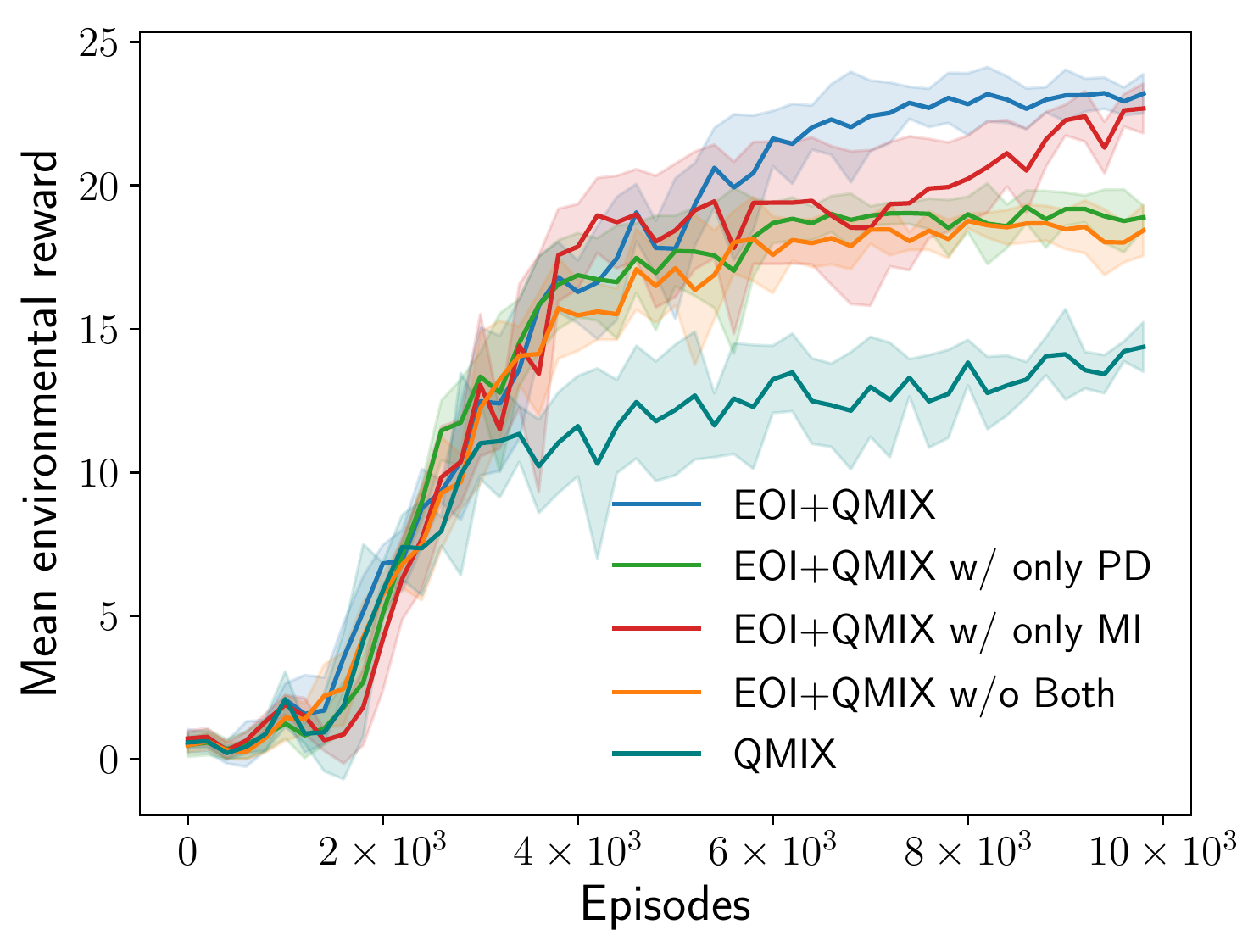}
	}

	\subfigure[EOI+MAAC]
	{
		\setlength{\abovecaptionskip}{3pt}
		\includegraphics[width=0.23\textwidth]{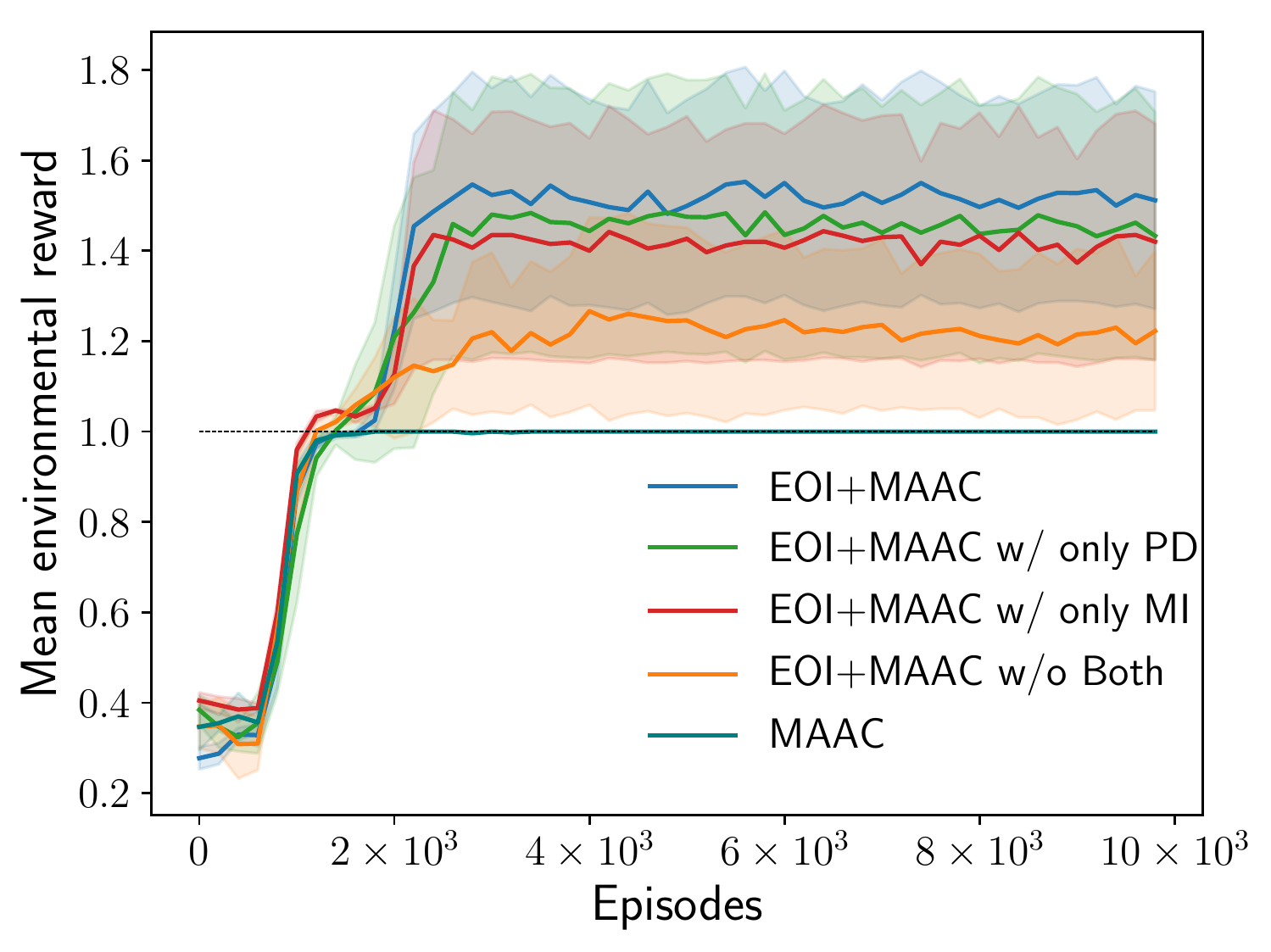}
	}
	\hspace{-0.60cm}
	\subfigure[EOI+QMIX]
	{
		\setlength{\abovecaptionskip}{3pt}
		\includegraphics[width=0.23\textwidth]{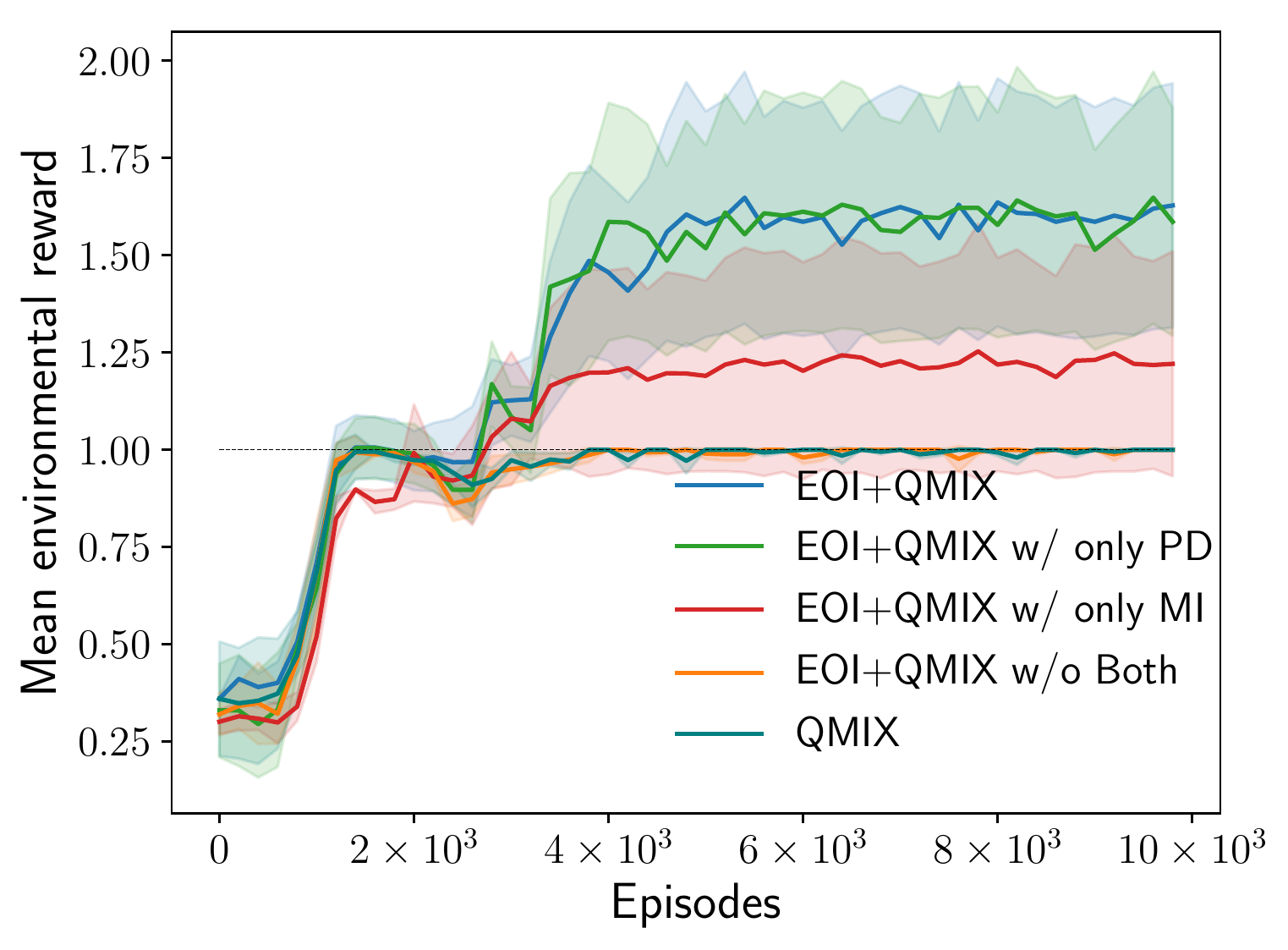}
	}
	\vspace{-0.20cm}
	\caption{Learning curves in Pac-Men (top) and Windy Maze (bottom).}
	\label{fig:curve_pm_wm}
\end{figure}

\begin{figure}[!t]
	\setlength{\abovecaptionskip}{3pt}
	\centering
	\subfigure[QMIX]
	{
		\setlength{\abovecaptionskip}{3pt}
		\includegraphics[width=0.22\textwidth]{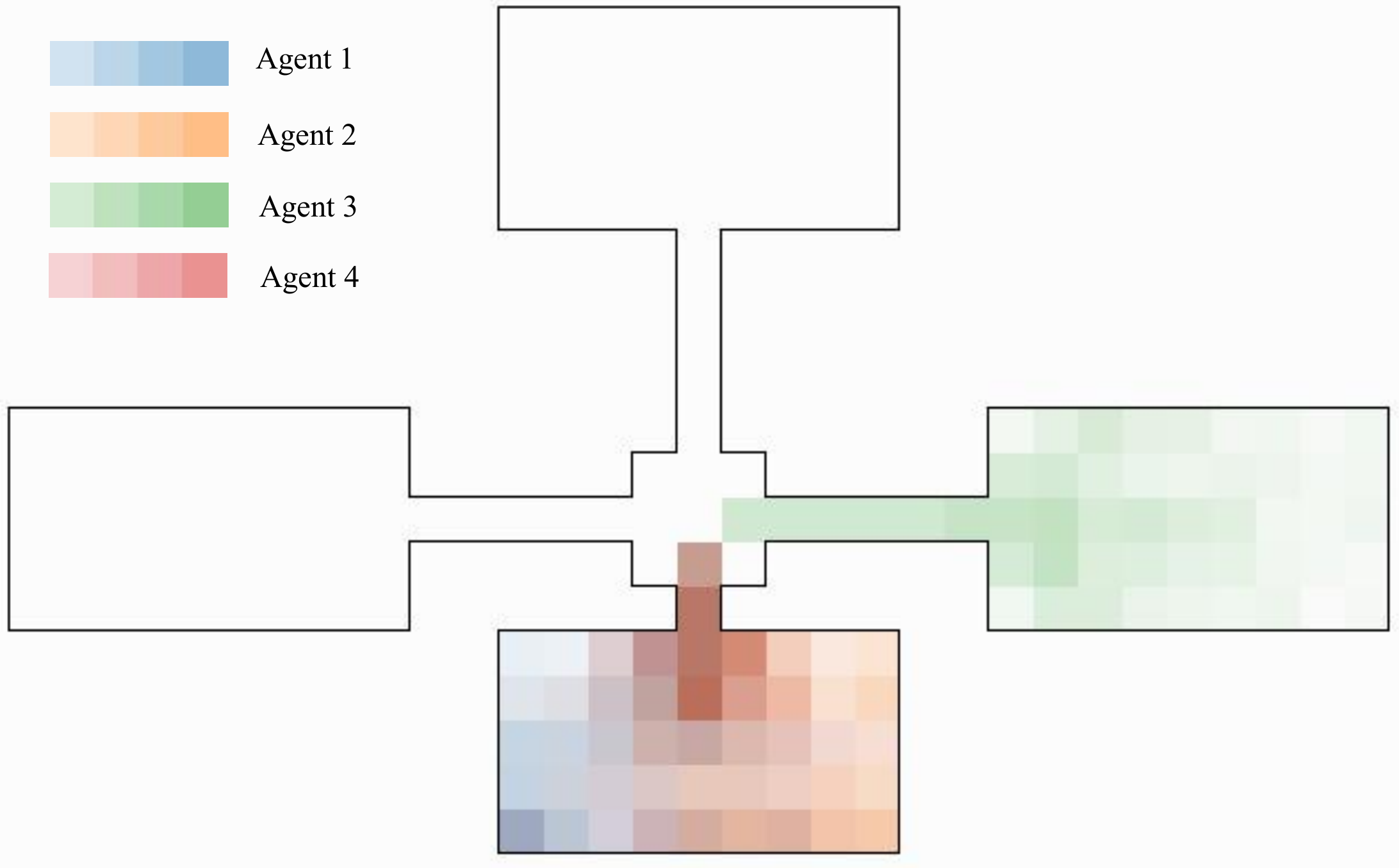}
		\label{fig:5_left} 
	}
	\hspace{-0.2cm}
	\subfigure[EOI+QMIX]
	{
		\setlength{\abovecaptionskip}{3pt}
		\includegraphics[width=0.22\textwidth]{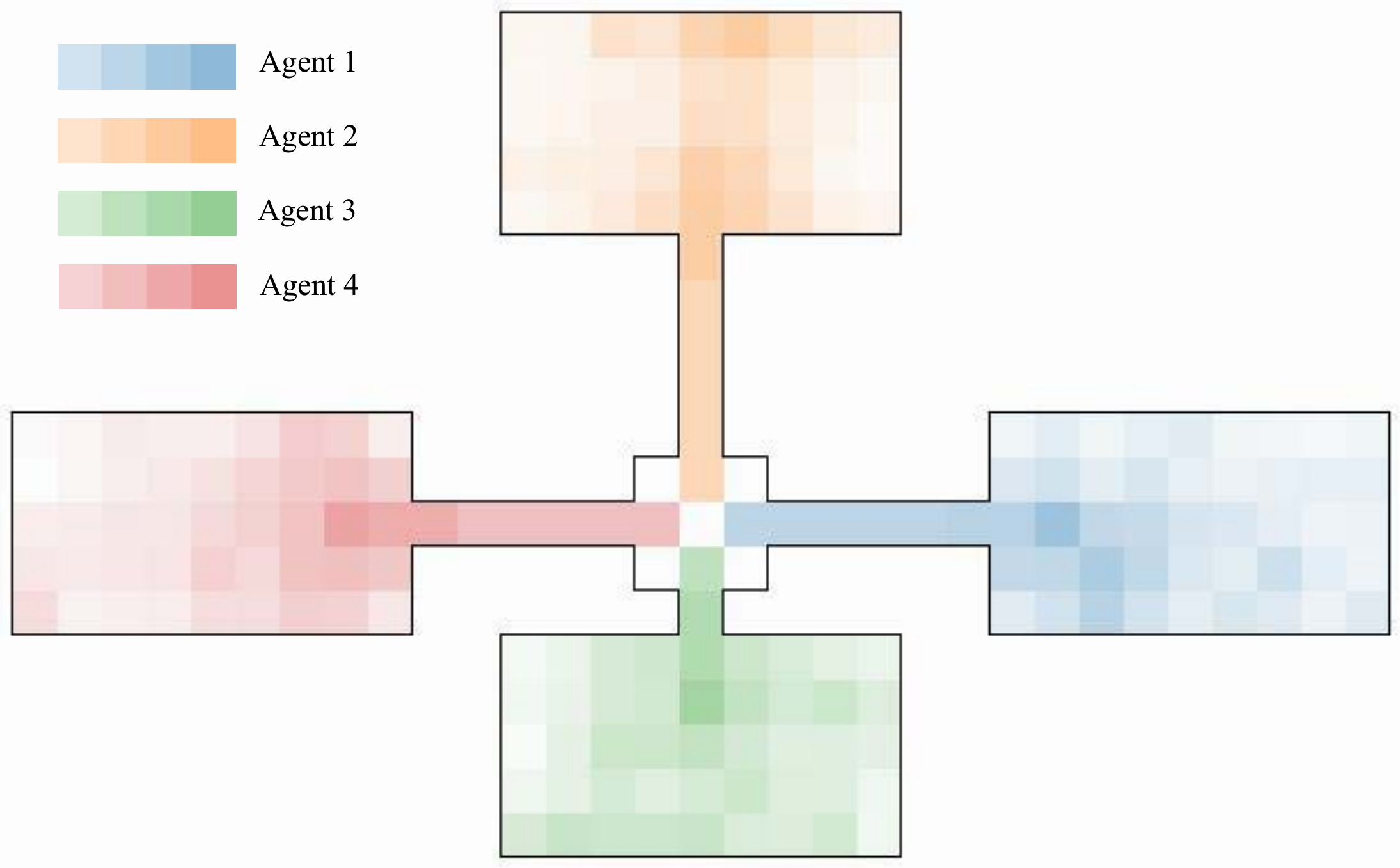}
		\label{fig:5_mid} 
	}
	\vspace{-0.2cm}
	\caption{Distributions of agents' positions of QMIX (a) and EOI+QMIX (b). The darker color means the higher value.}
	\label{fig:distribute_pm} 
\end{figure}

In Figure~\ref{fig:curve_pm_wm} (top), MAAC and QMIX get the lowest environmental reward respectively, since some agents learn to go to the same room and compete for the dots. This can be verified by the position distribution of QMIX agents in Figure~\ref{fig:5_left}, where three agents move to the same room. MAAC agents behave similarly. At the early training, it is easy for the agents to explore the bottom room and eat dots there to improve the environmental reward. Once the agents learn such policies, it is hard to explore other rooms, so the agents learn similar behaviors and fall at the local optimum.

\begin{figure}[t]
	\centering
	\includegraphics[width=0.2\textwidth]{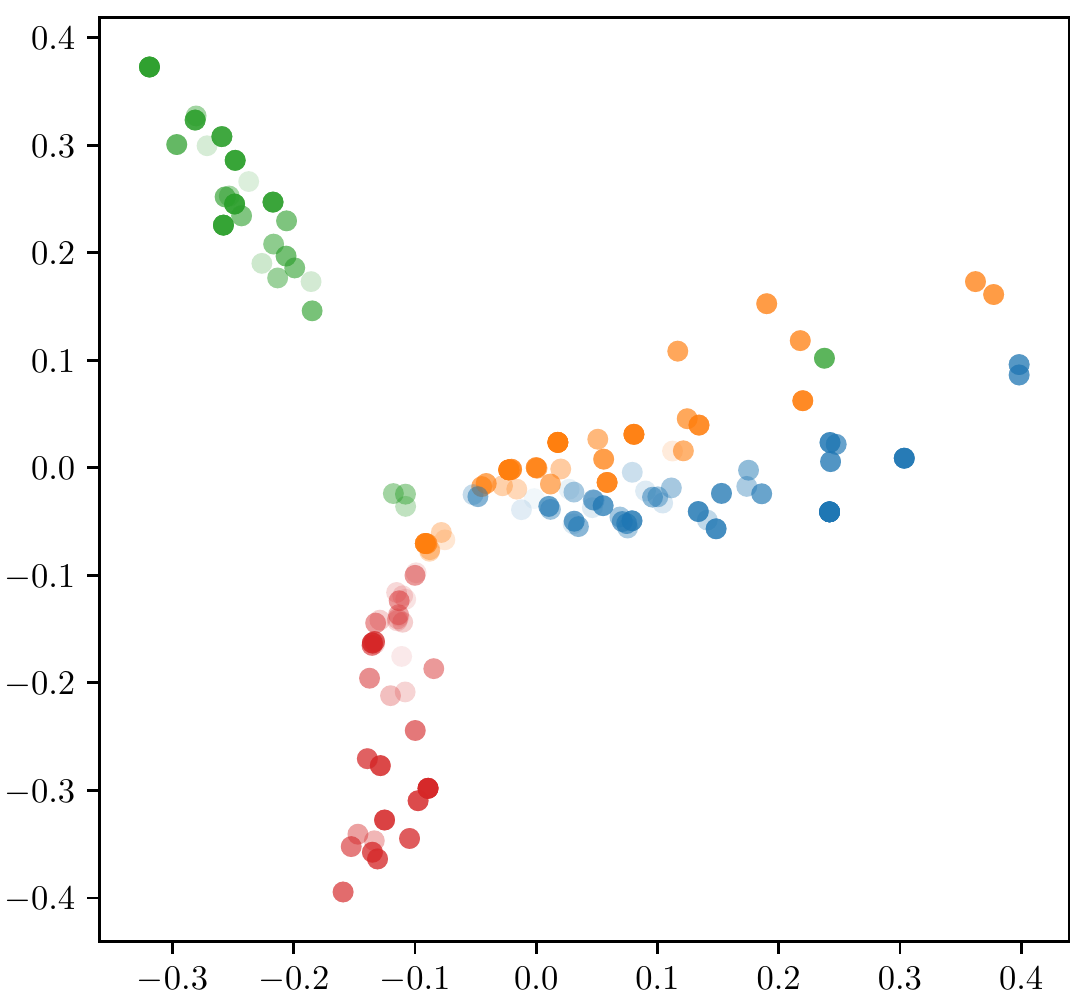}
	\vspace{-0.20cm}
	\caption{Kernel PCA of agents' observations of EOI+QMIX in Pac-Men. Darker color means higher correctly predicted probability of $P_{\phi}(I|O)$. }
	\label{fig:5_right} 
\end{figure}

Driven by the intrinsic reward without both regularizers, EOI obtains better performance than MAAC and QMIX. But the improvement is not significant since the observations of different agents cannot be easily distinguished when there is little difference between initial policies. The regularizers of PD and MI can increase the discriminability of $P_{\phi}(I|O)$, providing stronger intrinsic signals. Guided by $P_{\phi}(I|O)$ with PD or MI, the agents go to different rooms and eat more dots. MI theoretically increases the discriminability even the initial policies have no differences, while PD makes the observations distinguishable according to policies. Combining the advantages of the two regularizers leads to higher and steadier performance, as shown in Figure~\ref{fig:curve_pm_wm} (top). With both two regularizers, the agents respectively go to the four rooms and achieve the highest reward, which is indicated in Figure~\ref{fig:5_mid}. We also visualize the observations of different agents by kernel PCA, as illustrated in Figure~\ref{fig:5_right}, where darker color means higher correctly predicted probability of $P_{\phi}(I|O)$. We can see $P_{\phi}(I|O)$ can easily distinguish agents given their observations. 

\begin{figure}[t]
	\setlength{\abovecaptionskip}{3pt}
	\centering
	\subfigure[EOI+MAAC]
	{
		\setlength{\abovecaptionskip}{3pt}
		\includegraphics[width=0.23\textwidth]{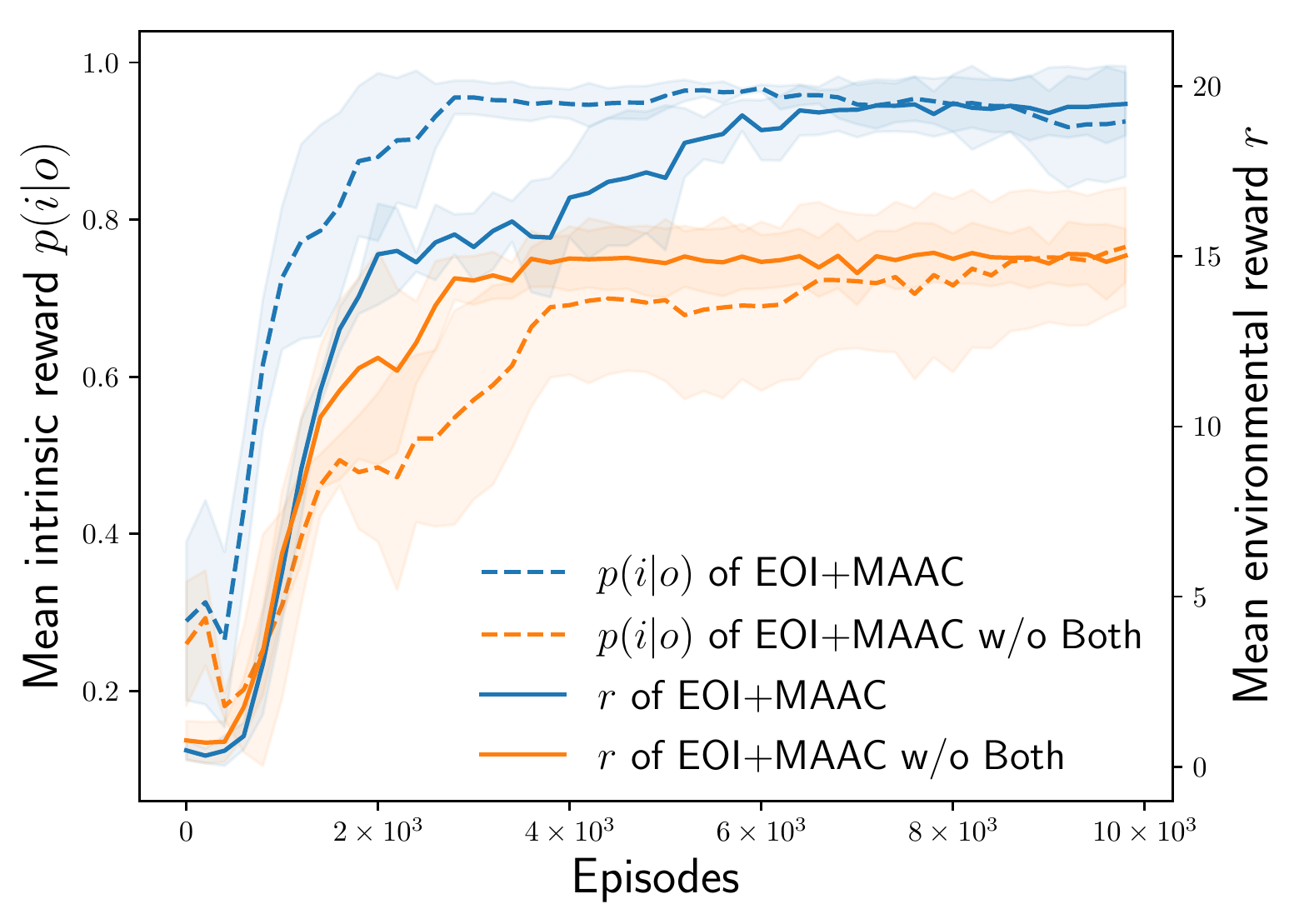}
	}
	\hspace{-0.4cm}
	\subfigure[EOI+QMIX]
	{
		\setlength{\abovecaptionskip}{3pt}
		\includegraphics[width=0.23\textwidth]{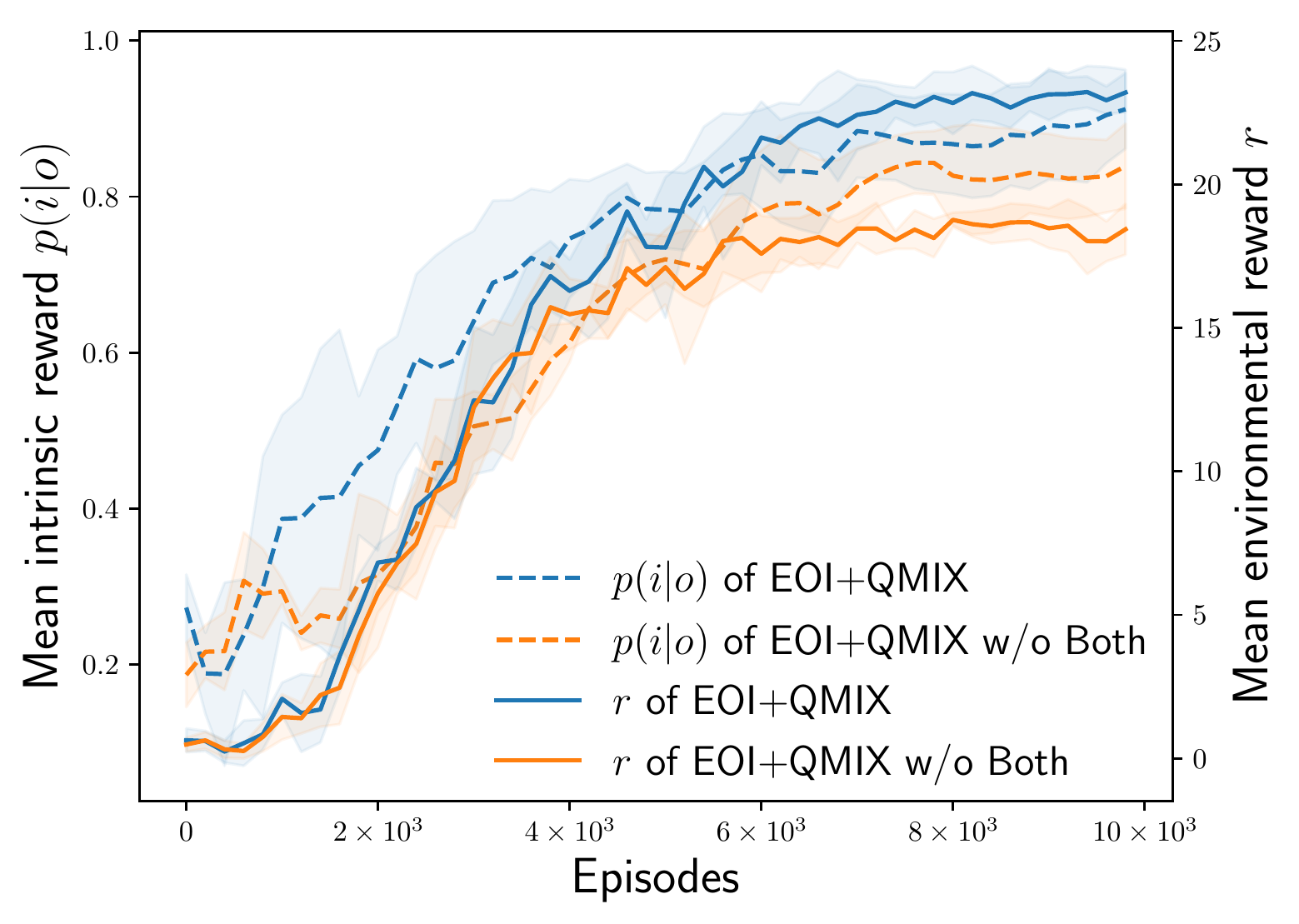}
	}
	\vspace{-0.20cm}
	\caption{Learning curves of intrinsic reward and environmental reward.}
	\label{fig:curve_pm_reward}
\end{figure}

\begin{figure}[!t]
	\setlength{\abovecaptionskip}{3pt}
	\centering
	\subfigure[Action distribution]
	{
		\setlength{\abovecaptionskip}{3pt}
		\includegraphics[width=0.19\textwidth]{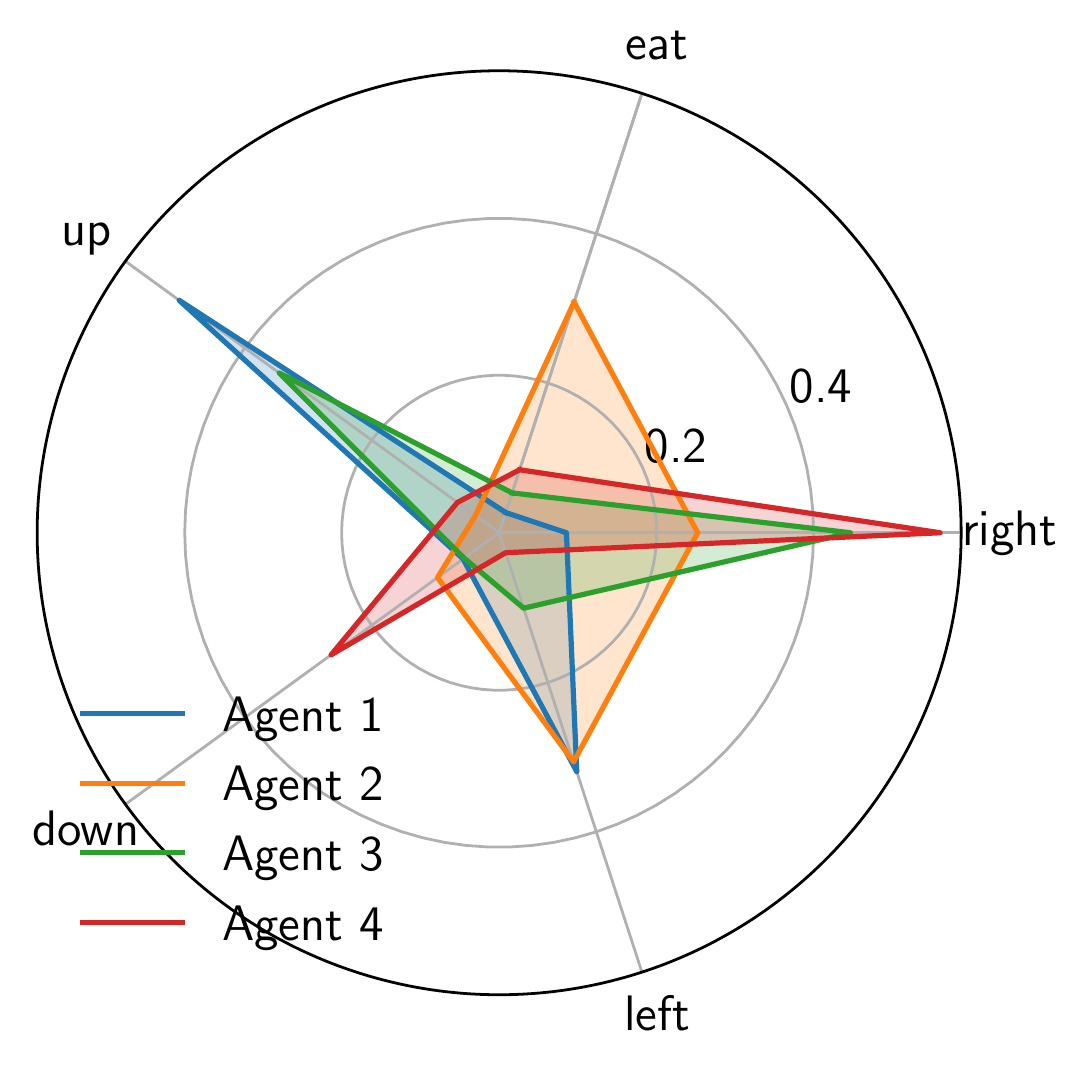}
		\label{fig:radar_initial}
	}
	\hspace{-0.20cm}
	\subfigure[EOI+QMIX]
	{
		\setlength{\abovecaptionskip}{3pt}
		\includegraphics[width=0.24\textwidth]{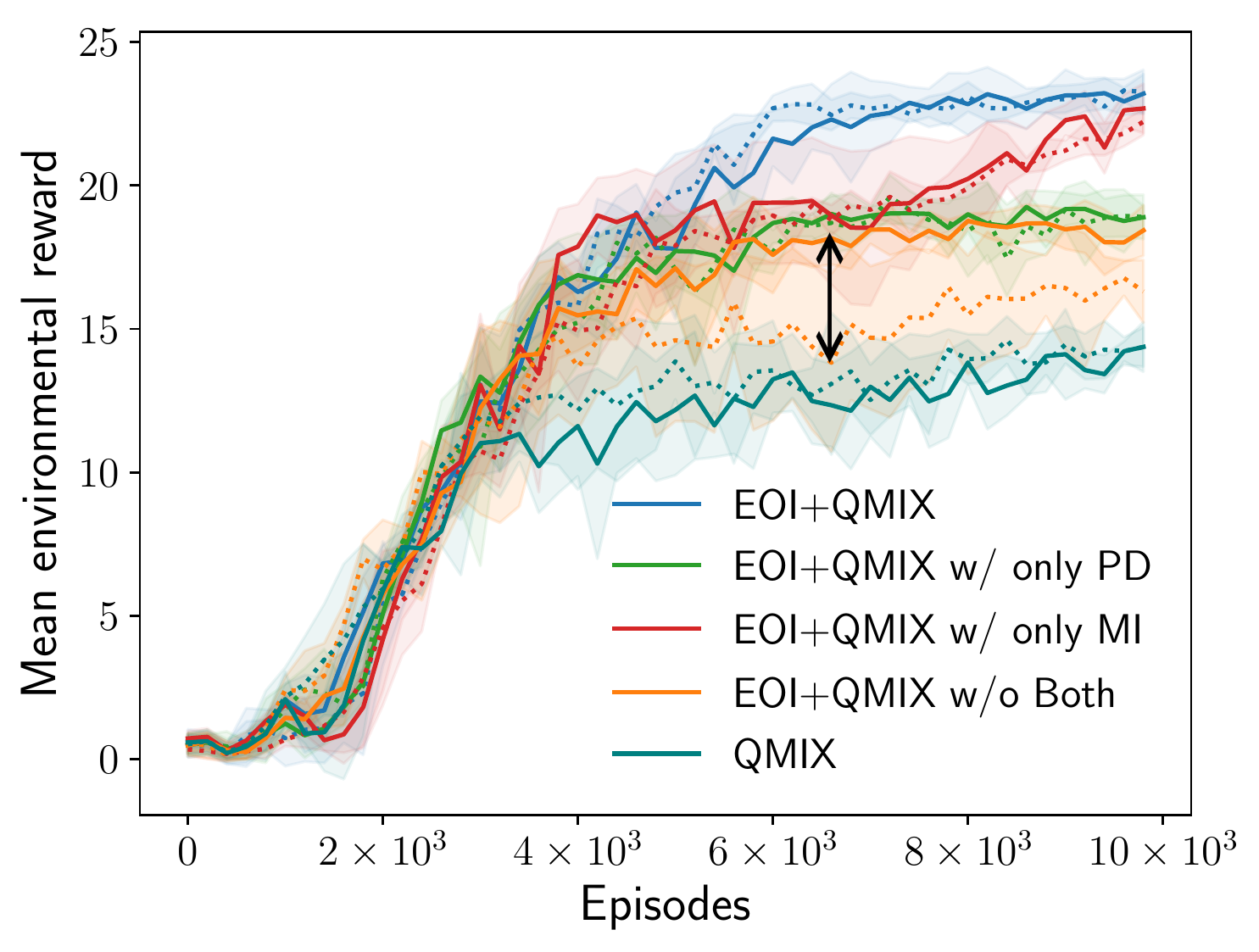}
		\label{fig:pm_initial}
	}
	\vspace*{-0.2cm}
	\caption{Action distributions of the initial action-value functions in QMIX and learning curves with EOI+QMIX. The dotted lines are the version with the same initial action-value functions.}
	\label{fig:initial}
\end{figure}

Similar results could be observed in Figure~\ref{fig:curve_pm_wm} (bottom). Under the effect of wind, it is easy to eat the right dot, even if the agent acts randomly. Limited by the time horizon, it is impossible that the agent first eats the right dot and then goes to the left end for the other dot. MAAC and QMIX only achieve the mean reward of $1$. Due to the wind and the small state space, the trajectories of the two agents are similar, thus EOI without regularizers provides little help, where the intrinsic reward is not strong enough to induce individuality. With regularizers, the observations on the right path and the left path can be discriminated gradually. In the learning process, the agents will first both learn to eat the right dot. Then one of them will change its policy and go left for higher intrinsic reward, and eventually the agents develop the distinct policies and get a higher mean reward.

To further investigate the effect of the regularizers, we show the learning curves of intrinsic reward and environmental reward of EOI with and without regularizers in Figure~\ref{fig:curve_pm_reward}. EOI with the regularizers converges to higher intrinsic reward than that without regularizers, meaning agents behave more distinctly. With regularizers, the rapid increase of intrinsic reward occurs before that of environmental reward, which indicates the regularizers make $P_{\phi}(I|O)$ also serve as the inductive bias for the emergence of individuality.

We also investigate the influence of the difference between initial policies. The action distributions (over visited observations) of the four initial action-value functions in QMIX are illustrated in Figure~\ref{fig:radar_initial}, where we find there is a large difference between them. The inherent difference makes agents distinguishable initially, which is the main reason EOI without the regularizers works well. We then initiate the four action-value functions with the same weights and re-run the experiments. The performance of EOI without the regularizers drops considerably, while EOI with the regularizers has almost no difference, as illustrated in Figure~\ref{fig:pm_initial}, indicating that PD and MI make the learning more robust to the initial policies. Agent individuality still emerges even with the same innate characteristics.

\begin{figure}[t]
	\setlength{\abovecaptionskip}{3pt}
	\centering
	\vspace{0.2cm}
	\includegraphics[width=0.26\textwidth]{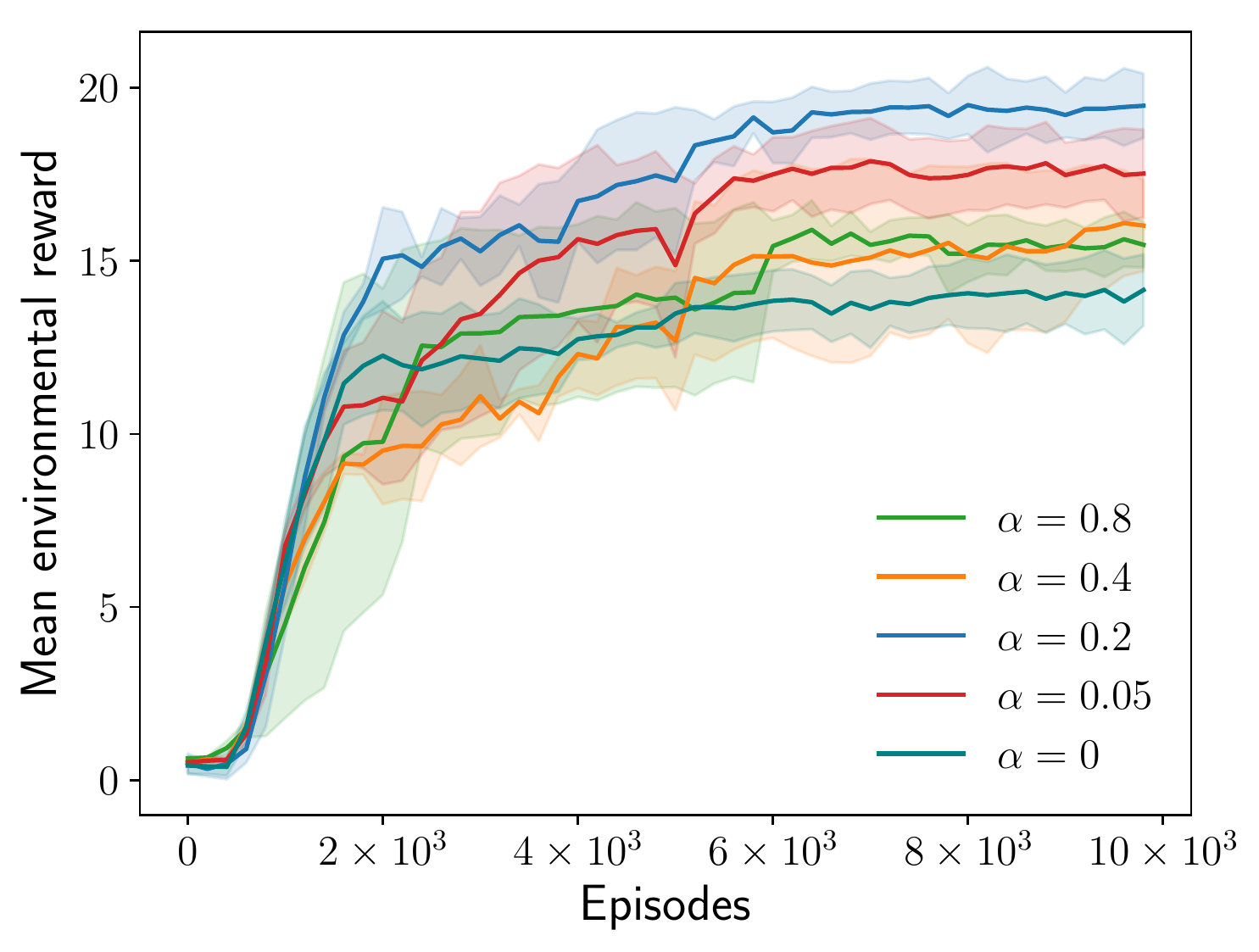}
	\vspace{-0.20cm}
	\caption{Learning curves of EOI+MAAC with different $\alpha$ in Pac-Men.}
	\label{fig:alpha}
	\vspace{-0.20cm}
\end{figure}

In Figure~\ref{fig:alpha}, we test EOI+MAAC with different $\alpha$ to investigate the effect of $\alpha$ on the emergent individuality. When $\alpha$ is too large, the agents will pay much attention to learn the individualized behaviors, which harms the optimization of the cumulative reward. The choice of $\alpha$ is a trade-off between success and individuality.

\subsection{Comparison with the Existing Methods}

\begin{figure}[t]
	\setlength{\abovecaptionskip}{3pt}
	\centering
	\subfigure[Firefighters]{
		\vspace{-1cm}
		\includegraphics[width=.222\textwidth]{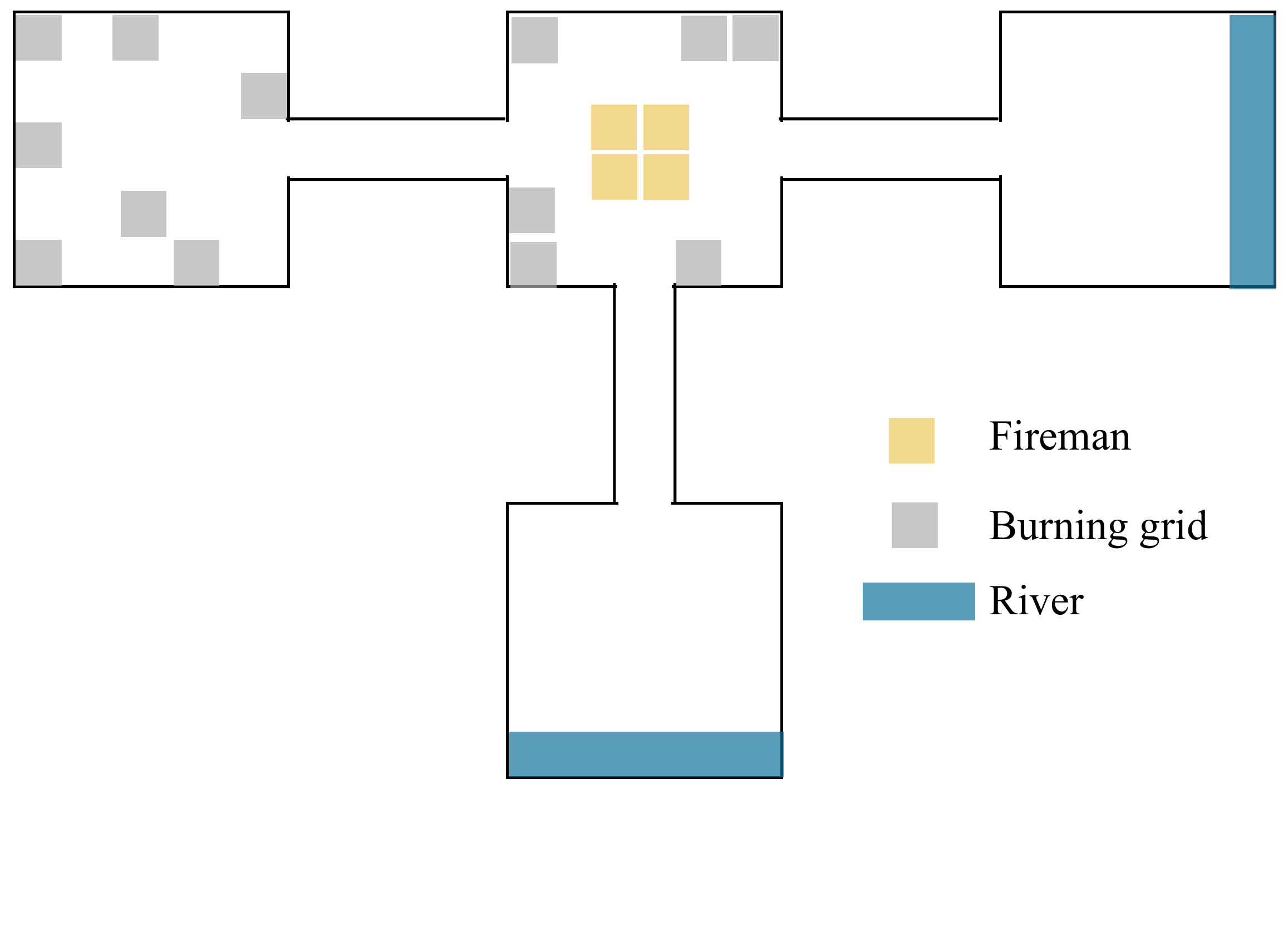}
		\label{fig:fm}
	}
	\hspace{-0.0cm}
	\subfigure[Learning curves]{
		\includegraphics[width=.229\textwidth]{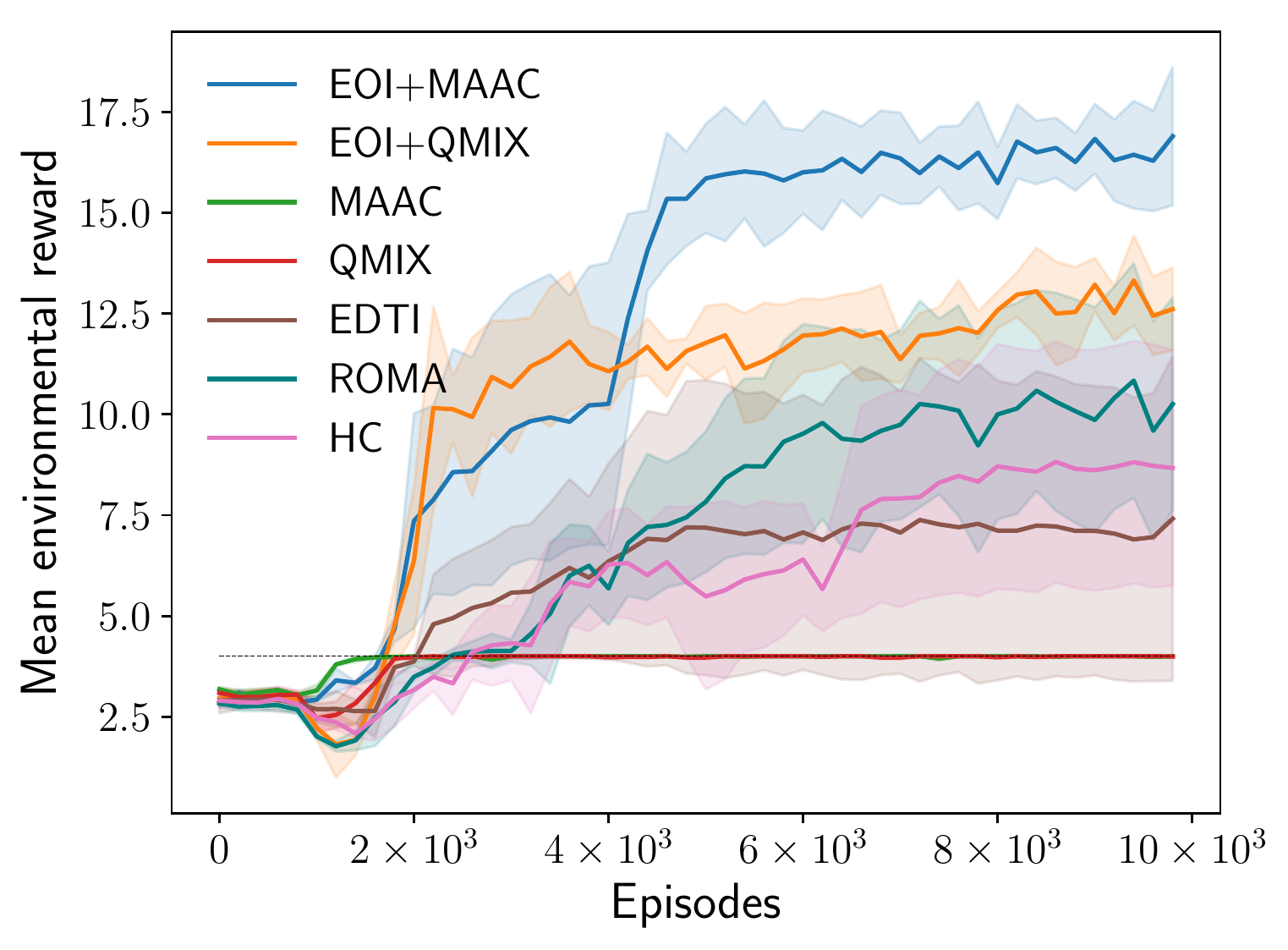}
		\label{fig:curve_fm}
	}
	\vspace{-0.2cm}
	\caption{Firefighters (a) and learning curves (b).}
	\vspace{-0.10cm}
\end{figure}

To compare EOI with previous methods, we design a more challenging task, Firefighters. There are some burning grids in two areas and two rivers in other areas. Four firefighters (agents) are initiated at the same burning area, illustrated in Figure~\ref{fig:fm}. The agent has a local observation that contains a square view with \(5\times5\) grids and can move to one of four neighboring grids, spray water, or pump water. They share a water tank, which is initiated with four units water. Once the agent sprays water on the burning grid, it puts out the fire and consumes one unit water. Once the agent pumps water in the river, the water in the tank increases by one unit. The agents only get a global reward, \textit{i.e.}, the number of the extinguished burning grids, at the final timestep.

As illustrated by Figure~\ref{fig:curve_fm}, MAAC and QMIX fall into local optimum. The agents learn to put out the fire around the initial position until the water is exhausted, and get the mean reward $4$. They do not take different roles of pumping water or putting out the fire farther. Benefited from the emergent individuality, EOI+MAAC achieves the best performance with a clear division of labor, where two agents go to the river for pumping water, and two agents go to different areas for fighting with fire. Since higher rewards are hard to explore, we compare EOI with EDTI, a multi-agent exploration method. EDTI gives the agent an intrinsic motivation, considering both curiosity and influence. It rewards the agent to explore the rarely visited states and to maximize the influence on the expected returns of other agents. EDTI could escape from the local optimum, but it converges to a lower mean reward than EOI. This is because encouraging curiosity and influence cannot help division of labor, and the computational complexity makes the learning inefficient especially in the environments with many agents. 

ROMA uses a role encoder taking the local observation to generate role embedding, and decodes the role to generate the parameters of individual action-value function in QMIX. However, ROMA converges slower than EOI+QMIX, since generating various parameters for the emergent roles is less efficient than encouraging diverse behaviors by reward shaping. Moveover, since the agents are initiated with the same observations, the role encoder will generate the similar role embeddings, which means similar agent behaviors at the beginning, bringing difficulty to the emergent roles.

HC first provides each agent a set of diverse skills, which are learned by DIAYN (before $6 \times 10^3$ episodes), then learns a high-level policy to select the skill for each agent. Since the skills are trained independently without centralized coordination, they could be diverse but might not be correlated with the success of task. So it is hard to explore and learn high-performing policies. Moreover, HC requires a centralized controller in execution. EOI encourages individuality with the coordination provided by centralized training, considering both of the global reward and other agents' policies. 

\begin{figure}[!t]
	\setlength{\abovecaptionskip}{3pt}
	\centering
	\subfigure[Battle]
	{
		\setlength{\abovecaptionskip}{3pt}
		\includegraphics[height=0.19\textwidth]{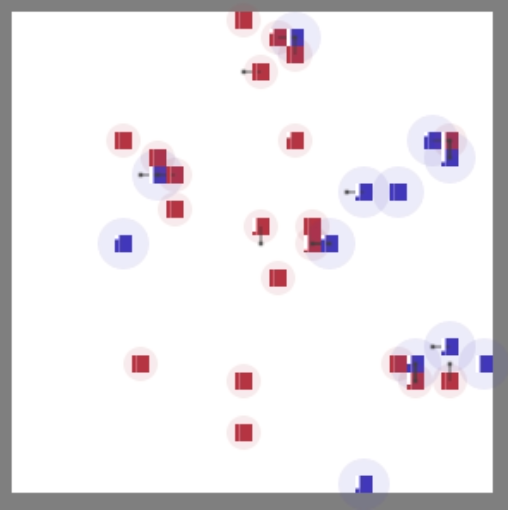}
	}
	\subfigure[10\_vs\_10]
	{
		\setlength{\abovecaptionskip}{3pt}
		\includegraphics[height=0.19\textwidth]{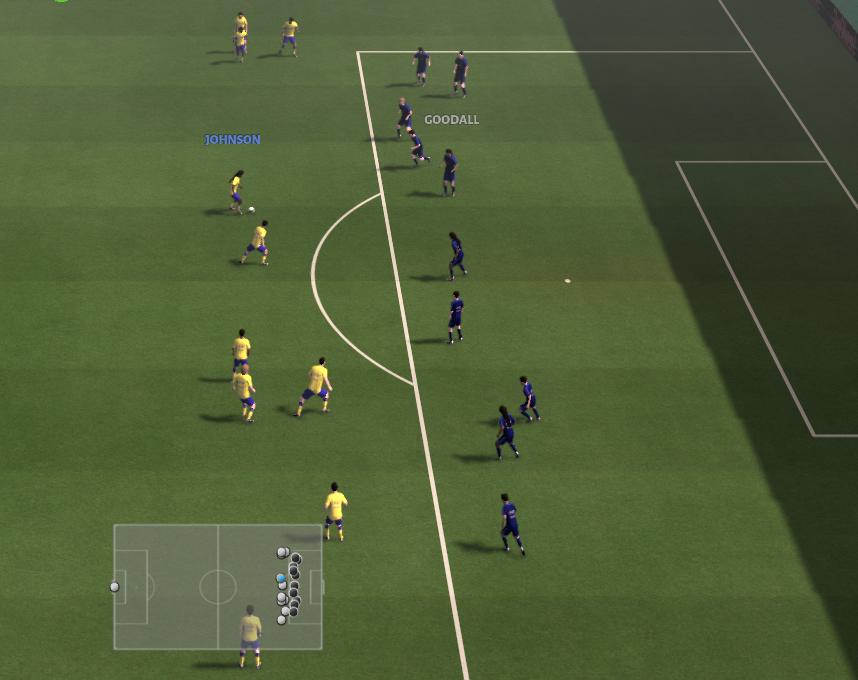}
	}
	\vspace{-0.2cm}
	\caption{Illustration of Battle and 10\_vs\_10.}
	\label{fig:Illustration_3}
	\vspace{-0.2cm}
\end{figure}

\begin{figure}[!t]
	\setlength{\abovecaptionskip}{3pt}
	\centering
	\subfigure[Battle]
	{
		\setlength{\abovecaptionskip}{3pt}
		\includegraphics[width=0.23\textwidth]{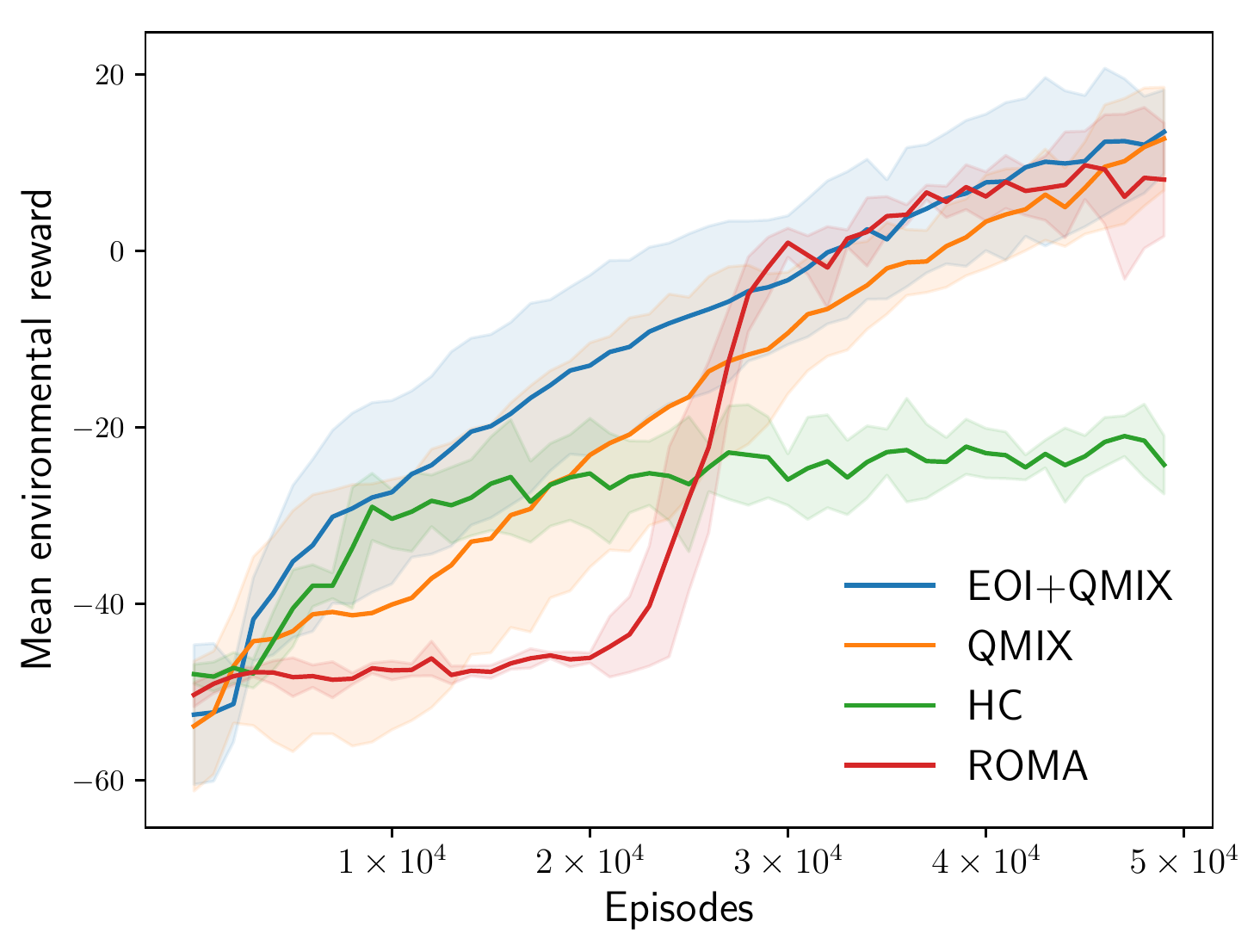}
	}
	\hspace{-0.4cm}
	\subfigure[10\_vs\_10]
	{
		\setlength{\abovecaptionskip}{3pt}
		\includegraphics[width=0.23\textwidth]{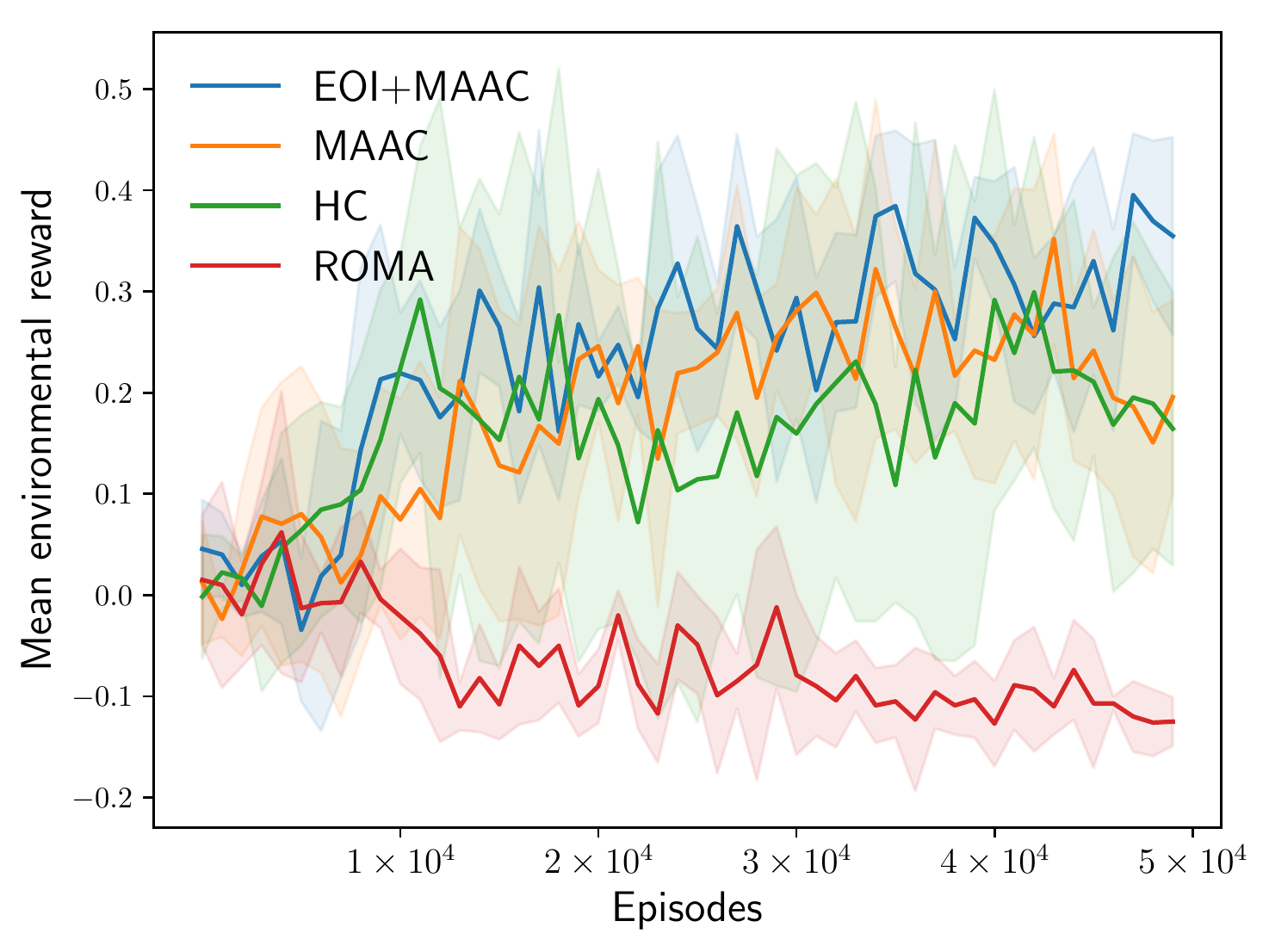}
	}
	\vspace*{-0.2cm}
	\caption{Learning curves in Battle and 10\_vs\_10.}
	\label{fig:large}
	\vspace*{-0.1cm}
\end{figure}

To investigate the effectiveness of EOI in large-scale complex environments, we test EOI in the two scenarios: MAgent \cite{zheng2018magent} task Battle and gfootball \cite{kurach2020google} task 10\_vs\_10, as illustrated in Figure~\ref{fig:Illustration_3}.
\begin{itemize}
	\vspace{-0.2cm}
	\item \textit{Battle.} 20 agents (red) battle with 12 enemies (blue) which are controlled by built-in AI. The moving or attacking range of the agent is four neighbor grids. If an agent attacks an enemy, kills an enemy, and is killed by an enemy, it will respectively get a reward of $+1$, $+10$, and $-2$. The team reward is the sum of all agents' rewards.
	\item \textit{10\_vs\_10.} 10 agents try to score facing 10 defenders which are controlled by built-in AI. If the agents shoot the ball into the goal, they will get a global reward of $+1$. If they lose control of the ball, they will get a global reward of $-0.2$.
	\vspace{-0.2cm}
\end{itemize}

\begin{figure*}[!t]
	\setlength{\abovecaptionskip}{3pt}
	\centering
	\subfigure[so\_many\_baneling]
	{
		\setlength{\abovecaptionskip}{3pt}
		\includegraphics[width=0.24\textwidth]{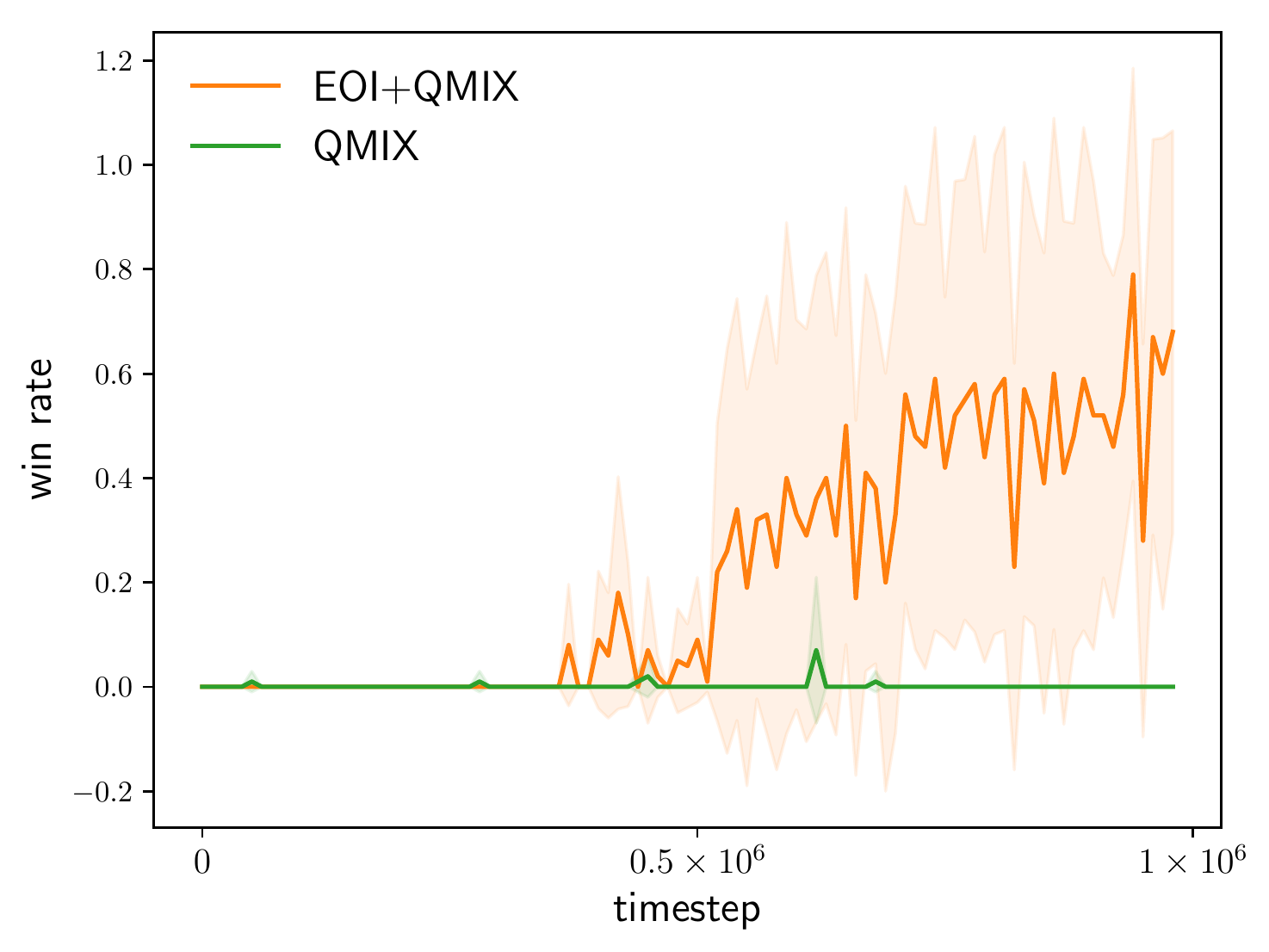}
	}
	\hspace{-0.3cm}
	\subfigure[2c\_vs\_64zg]
	{
		\setlength{\abovecaptionskip}{3pt}
		\includegraphics[width=0.24\textwidth]{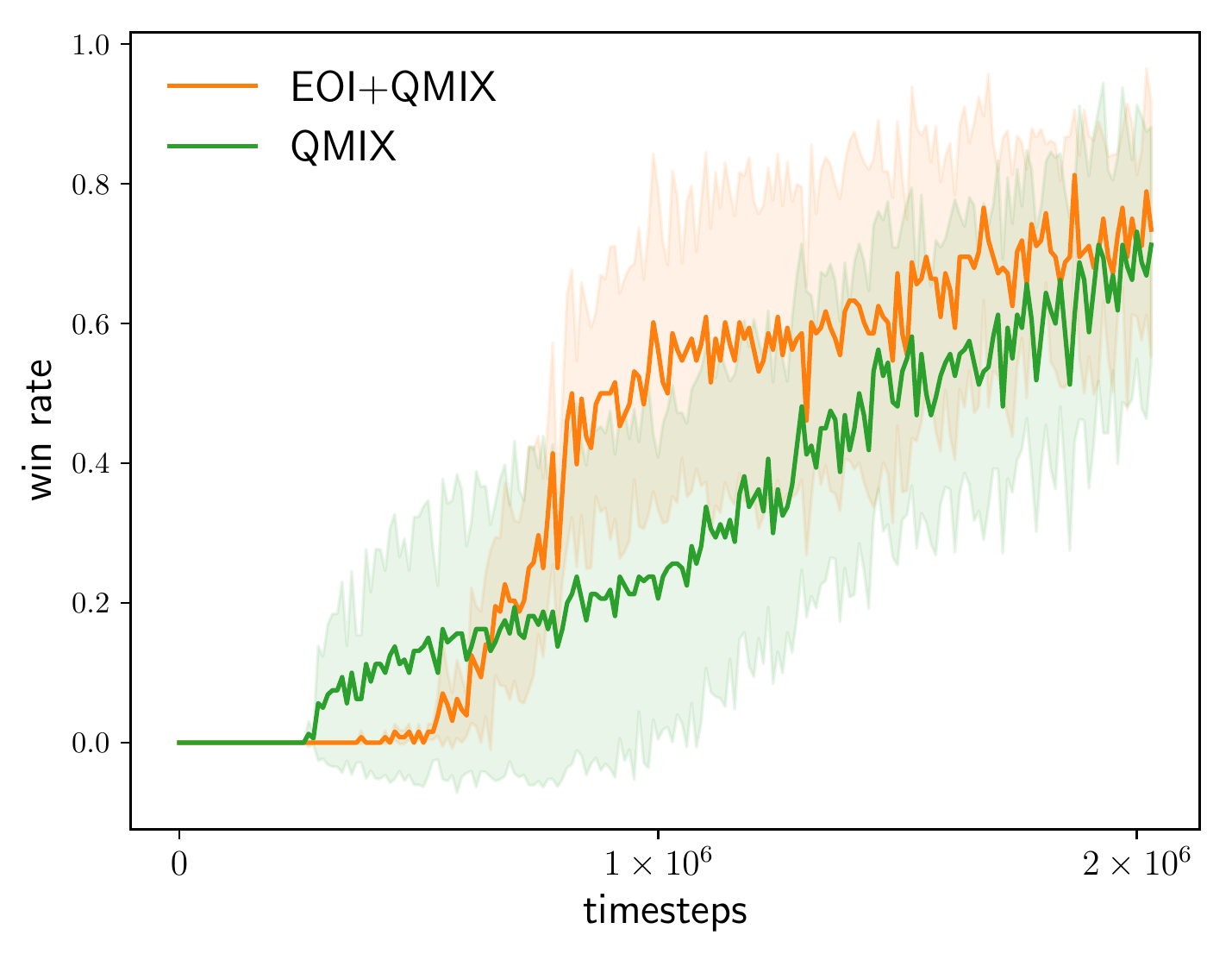}
	}
	\hspace{-0.3cm}
	\subfigure[3s\_vs\_5z]
	{
		\setlength{\abovecaptionskip}{3pt}
		\includegraphics[width=0.24\textwidth]{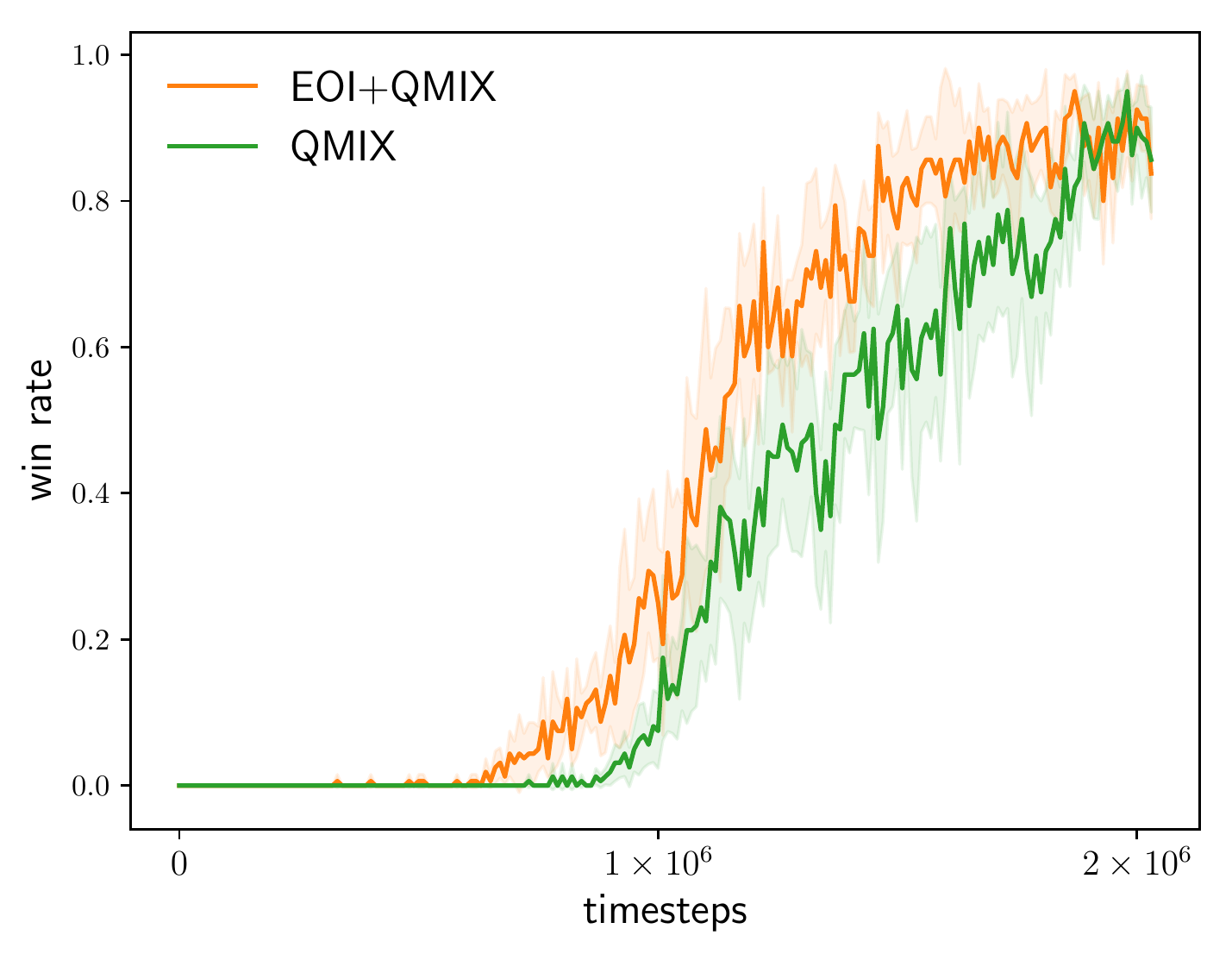}
	}
	\hspace{-0.3cm}
	\subfigure[5m\_vs\_6m]
	{
		\setlength{\abovecaptionskip}{3pt}
		\includegraphics[width=0.24\textwidth]{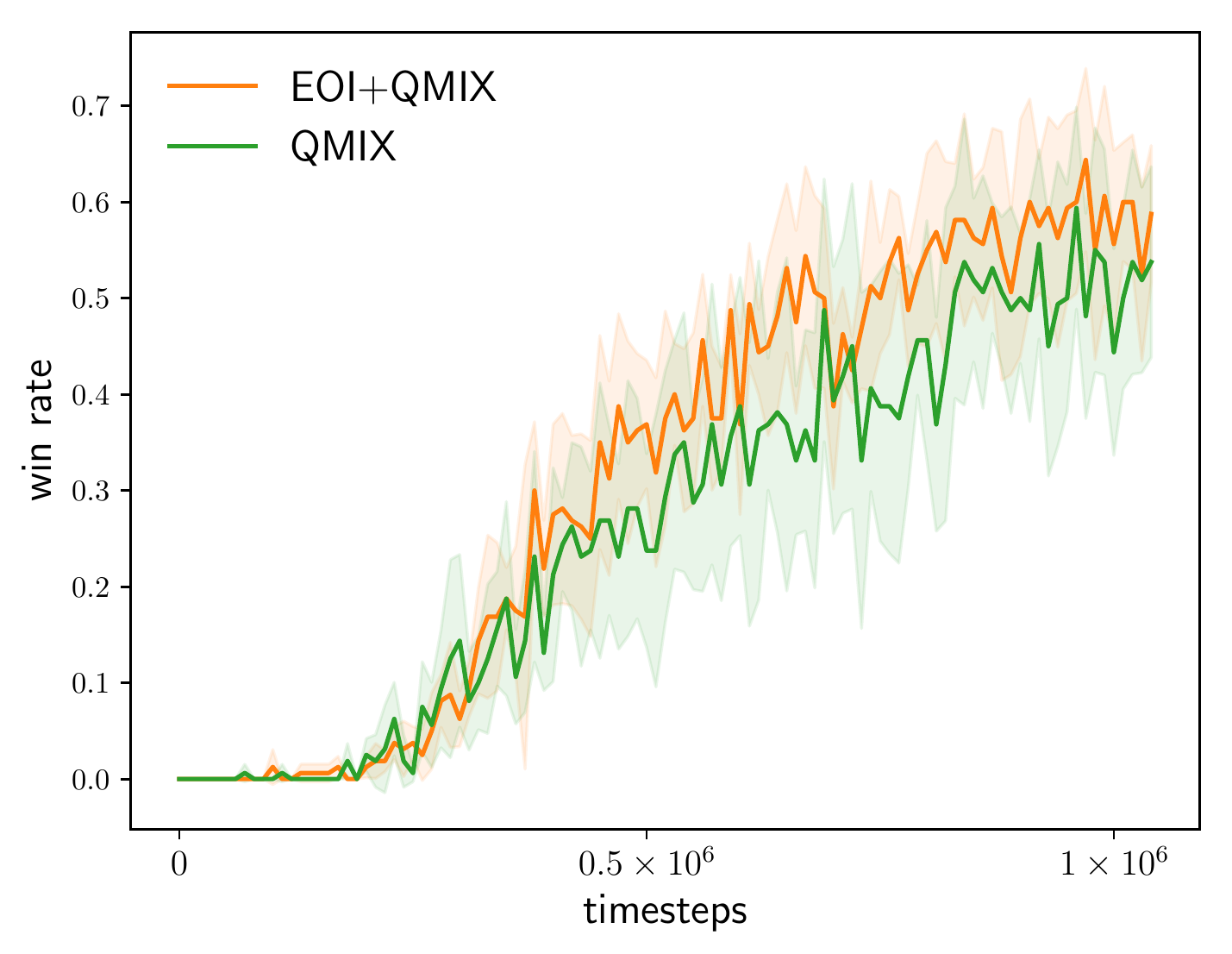}
	}
	\vspace*{-0.4cm}
	\caption{Learning curves in SMAC tasks.}
	\label{fig:smac}
	\vspace*{-0.3cm}
\end{figure*}

Due to the large number of agents, we let the agents share the weights of neural network $\theta$ and additionally feed a one-hot encoding of agent index into $\theta$. The classifier $\phi$ does not use the index information to avoid overfitting. Parameter sharing might cause similar behaviors, which hinders the emergence of complex cooperative strategies. However, EOI helps a number of agents learn individualized policies even with shared parameters. As shown in Figure~\ref{fig:large}, EOI increases the behavior diversification and outperforms the baselines. Moreover, we find that the coefficient $\alpha$ should be small in these environments where the success of the task is not always aligned with individuality. When individuality conflicts with success, focusing on individuality would negatively impact the maximization of environmental rewards. Hence we linearly decrease $\alpha$ to zero during the training, which could accelerate the emergence of individuality at the early training and make the optimization objective unbiased at the later training. It is beneficial to explore the mechanism for adaptive $\alpha$ to balance the success and individuality, \textit{e.g.}, to adjust $\alpha$ by meta gradient with the direction of the fastest increase of the environmental reward as in \citet{lin2019adaptive}. 

Then we test EOI on SMAC tasks \cite{samvelyan19smac}, and the results are presented in Figure~\ref{fig:smac}. So\_many\_baneling is a task where 7 Zealots fight with 32 Banelings. The key to winning this task is that the Zealots should cooperatively spread out around the map far from each other so that the Banelings' damage is distributed as thinly as possible. However, the original so\_many\_baneling is too easy. We set reward\_sparse=True, sight\_range=4.0, shoot\_range=0.0, move\_amount=0.3, and simplify the observed information. The modified version is much more difficult, and vanilla QMIX is hard to explore the winning experiences. EOI learns individualized strategies which are crucial to solve this task. We adopt the default settings of other three hard maps: 2c\_vs\_64zg, 3s\_vs\_5z, and 5m\_vs\_6m. 2c\_vs\_64zg requires positioning micro-trick and 3s\_vs\_5z requires kiting micro-trick \cite{samvelyan19smac}, thus EOI obtains significant performance gain in the two tasks. Moreover, empirically we find that the agents are more likely to benefit from individualized behaviors if the trajectory is longer. The code of EOI on SAMC tasks is publisded at \url{https://github.com/jiechuanjiang/EOI_on_SMAC} and \url{https://github.com/jiechuanjiang/eoi_pymarl}.

\subsection{Similarity and Difference Between EOI and DIAYN}

DIAYN is proposed to learn diverse skills with single-agent RL in the absence of any rewards. It trains the agent by maximizing the mutual information between skills ($Z$) and states ($S$), maximizing the policy entropy, and minimizing the mutual information between skills and actions ($A$) given the state. The optimization objective is
\begin{equation}
\nonumber
\begin{split}
&\mathrm{MI}(S; Z)+\mathcal{H}(A | S)-\mathrm{MI}(A ; Z | S)\\
&=\mathcal{H}(Z)-\mathcal{H}(Z|S)+\mathcal{H}(A|S)-\left(\mathcal{H}(A | S)-\mathcal{H}(A | Z, S)\right)\\
&=\mathcal{H}(Z)-\mathcal{H}(Z | S)+\mathcal{H}(A | Z, S).
\end{split}
\end{equation}
To maximize this objective, in practice DIAYN gives the learning agent an intrinsic reward $\log q(z|s)-\log p(z)$, where $q(z|s)$ approximates $p(z|s)$ and $p(z)$ is a prior distribution. EOI gives each agent an intrinsic reward of $p(i|o)$. Let agents correspond to skills and observations correspond to states, then EOI has a similar intrinsic reward with DIAYN, though the motivations are distinct. Unlike from DIAYN, EOI employs two regularizers to capture the unique characteristics of MARL, strength the reward signals, and promote the emergence of individuality in MARL.  

\begin{figure}[t]
	\centering
	\includegraphics[width=0.25\textwidth]{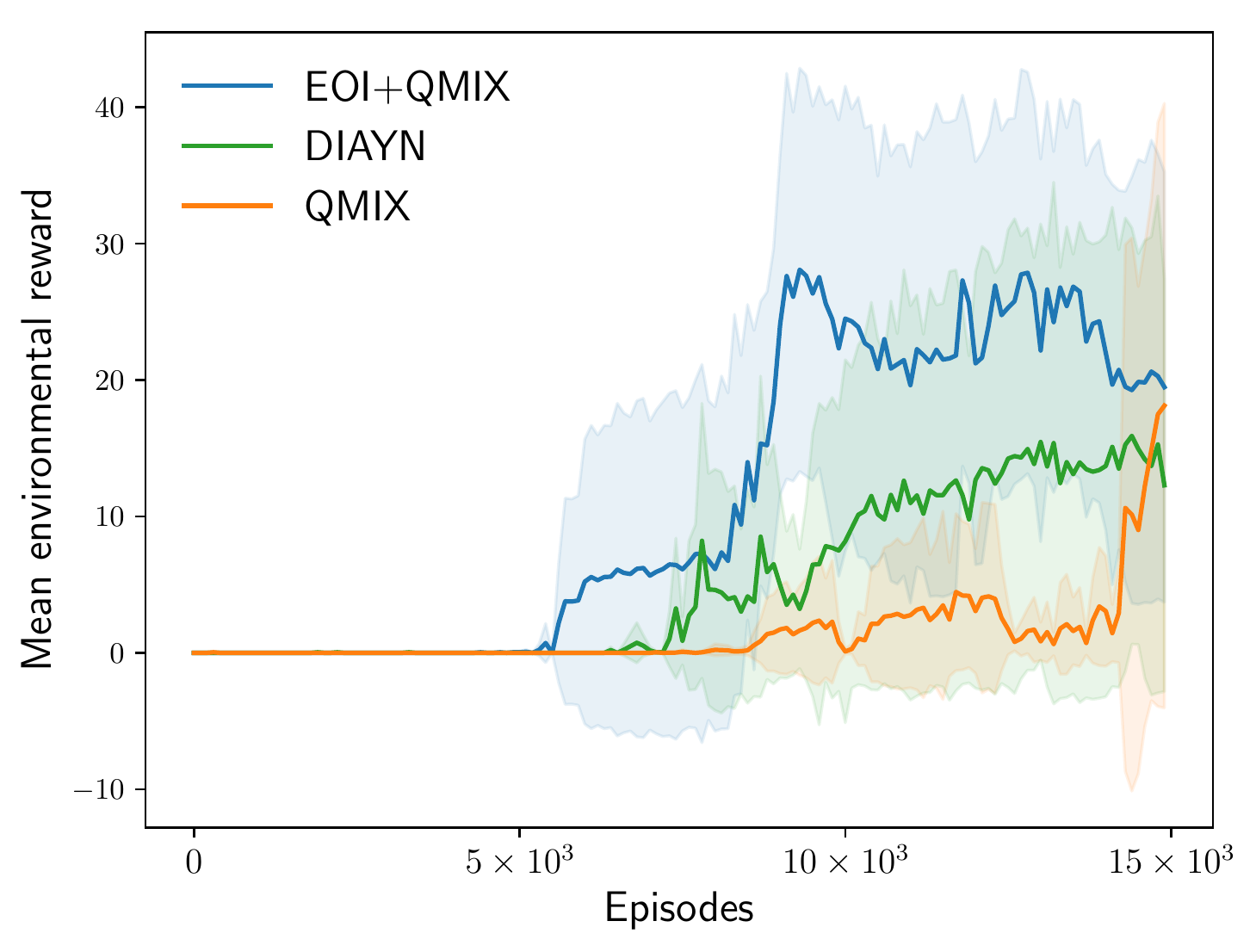}
	\vspace{-0.50cm}
	\caption{Learning curves in sparse Pac-Men.}
	\label{fig:sparse_pm}
	\vspace{-0.50cm}
\end{figure}

To empirically investigate the difference between EOI and the original DIAYN (making each skill as an agent), we test them in a sparse-reward version of Pac-Men, where the reward is defined as the minimum eaten dots of the four rooms. The results are shown in Figure~\ref{fig:sparse_pm}. In this sparse-reward task, the agents could hardly obtain environmental reward before the individuality emerges. Without encouraging individuality, QMIX learns slowly. EOI+QMIX explores the environmental reward earlier and converges faster than DIAYN, indicating EOI has a stronger ability for individuality.

\section{Conclusion and Discussion}

We have proposed EOI, a novel method for the emergence of individuality in MARL. EOI learns a probabilistic classifier that predicts a probability distribution over agents given their observation and gives each agent an intrinsic reward of being correctly predicted by the classifier. Two regularizers are introduced to increase the discriminability of the classifier. We realized EOI on top of two popular MARL methods and empirically demonstrated that EOI outperforms existing methods in a variety of multi-agent cooperative tasks. We also discuss the similarity and difference between EOI and DIAYN. 

However, EOI might be limited in some scenarios where the observation or trajectory cannot represent individuality. For example, if the agents could get the same full observation of the state, the probabilistic classifier cannot discriminate different agents based on the same observation. Moreover, if the local observation contains the identity information, \textit{e.g.}, agent index, the probabilistic classifier would overfit the identity information and cannot help the emergence of individuality. We leave the limitations to future work.

\section*{Acknowledgements}

This work is supported by NSF China under grant 61872009.

\nocite{langley00}

\bibliography{ref}
\bibliographystyle{icml2021}

\clearpage
\onecolumn
\appendix

\section{Mathematical Formulation}
\label{app:bilevel}

Let $\boldsymbol{\theta}$ and $\phi$ denote the parameters of the joint policies and the probabilistic classifier, respectively. Then, the whole learning process corresponds to the following bi-level optimization:
\begin{align*}
\nonumber
\max_{\boldsymbol{\theta} \in \Theta} & \quad J(\boldsymbol{\theta}, \phi^*(\boldsymbol{\theta})) \\
\nonumber s.t.  &  \quad \phi^*(\boldsymbol{\theta})=\arg \underset{\phi' \in \Phi}{\min} \mathcal{L}(\phi', \boldsymbol{\theta}),
\end{align*} 
where $J$ is the RL objective with intrinsic reward, $\mathcal{L}$ is the loss function of the probabilistic classifier, and $\phi$ is an implicit function of $\boldsymbol{\theta}$. Therefore, to solve this optimization, we can iteratively update $\boldsymbol{\theta}$ by
\begin{equation}
\nonumber
\frac{\mathrm{d} J(\boldsymbol{\theta}, \phi^*(\boldsymbol{\theta}))}{\mathrm{d} \boldsymbol{\theta}} = \frac{\partial J(\boldsymbol{\theta}, \phi)}{\partial \boldsymbol{\theta}}\bigg\rvert_{\phi=\phi^*(\boldsymbol{\theta})} + \frac{\mathrm{d} \phi^*(\boldsymbol{\theta}) }{ d \boldsymbol{\theta}} \frac{\partial J(\boldsymbol{\theta}, \phi)}{\partial \phi}\bigg\rvert_{\phi=\phi^*(\boldsymbol{\theta})}
\end{equation}
where  
\begin{equation}
\nonumber
\frac{\mathrm{d} \phi^*(\boldsymbol{\theta}) }{ d \boldsymbol{\theta}} = - \left(  \frac{\partial^2 \mathcal{L}(\phi, \boldsymbol{\theta})} {\partial\phi \partial \phi^T} \right)^{-1} \left(  \frac{\partial^2 \mathcal{L}(\phi, \boldsymbol{\theta})} {\partial\phi \partial\boldsymbol{\theta}^T} \right) \bigg\rvert_{\phi=\phi^*(\boldsymbol{\theta})}
\end{equation}
which is obtained by the implicit function theorem. In practice, the second-order term is neglected due to high computational complexity, without incurring significant performance drop, such as in meta-learning and GANs. Therefore,  we can solve the bi-level optimization by the first-order approximation with iterative updates:
\begin{align*}
\nonumber
\phi_{k+1}  &\approx \arg \min_\phi \mathcal{L}(\phi, \mathcal{B}_k) \\
\nonumber
\boldsymbol{\theta}_{k+1}  &=  \boldsymbol{\theta}_k + \zeta_k \nabla_{\boldsymbol{\theta}}J(\boldsymbol{\theta}, \phi_{k+1}). 
\end{align*}

%
%

\section{Hyperparameters}
\label{app:hyper}
The hyperparameters of EOI and the baselines in each scenario are summarized in Table~\ref{tab:hyperparameter}. Since QMIX and MAAC are off-policy algorithms with replay buffer, we do not need to maintain the buffer $\mathcal{B}$ but build the training data from the replay buffer $\mathcal{D}$. For EDTI, ROMA, and HC, we use their default settings.

\begin{table*}[h]
	\setlength{\abovecaptionskip}{3pt}
	\renewcommand{\arraystretch}{0.9}
	\centering
	\caption{Hyperparameters}
	\vskip 0.1cm
	\label{tab:hyperparameter}
	\begin{footnotesize}
		\begin{tabular}{@{}cccccc@{}}
			\toprule
			Hyperparameter&Pac-man&Windy Maze&Firefighters&Battle&10\_vs\_10\\
			\midrule
			runs with different seeds&$5$&$10$&$5$&$5$&$5$\\
			horizon ($T$) &$30$&$15$&$20$&$100$&$100$\\
			discount (\(\gamma\)) &\multicolumn{3}{c}{$0.98$}&$0.96$&$0.995$\\
			replay buffer size &\multicolumn{4}{c}{$2 \times 10^{4}$}&$1 \times 10^{4}$ \\
			actor learning rate &\multicolumn{3}{c}{$1 \times 10^{-3}$}&-&$3 \times 10^{-4}$\\
			critic learning rate &\multicolumn{3}{c}{$1 \times 10^{-4}$}&-&$1 \times 10^{-4}$\\
			QMIX learning rate &\multicolumn{4}{c}{$1 \times 10^{-4}$}&-\\
			\# MLP units &\multicolumn{5}{c}{$(128,128)$}\\
			batch size &\multicolumn{5}{c}{$128$} \\
			MLP activation &\multicolumn{5}{c}{ReLU}\\
			optimizer&\multicolumn{5}{c}{Adam}\\
			
			\midrule
			$\phi$ learning rate &\multicolumn{3}{c}{$1 \times 10^{-3}$}&$1 \times 10^{-4}$&$1 \times 10^{-4}$\\
			$\alpha$ in QMIX &\multicolumn{3}{c}{$0.05$}&$0.02$&-\\
			$\alpha$ in MAAC &\multicolumn{3}{c}{$0.2$}&-&$0.04$\\
			$\beta_1$  &\multicolumn{3}{c}{$0.04$}&$0.05$&$0.05$\\
			$\beta_2$  &\multicolumn{3}{c}{$0.1$}&$0.05$&$0.05$\\
			$\Delta t$  &\multicolumn{5}{c}{$4$}\\
			
			\bottomrule
		\end{tabular}
	\end{footnotesize}
\end{table*}

\end{document}